\documentclass[journal]{IEEEtran}
\usepackage{amsmath,amsfonts}
\usepackage{algpseudocode}
\usepackage{algorithm}
\usepackage{array}
\usepackage[caption=false,font=normalsize,labelfont=sf,textfont=sf]{subfig}
\usepackage{textcomp}
\usepackage{stfloats}
\usepackage{url}
\usepackage{verbatim}
\usepackage{graphicx}
\usepackage{booktabs}
\usepackage{multirow}
\usepackage{cite}
\usepackage{hyperref}
\usepackage{enumitem}

\newcommand{\sysName}{AACD}

\begin{document}

\title{TacSL: A Library for Visuotactile Sensor Simulation and Learning}

\author{Iretiayo Akinola$^1$$^*$, Jie Xu$^1$$^*$, Jan Carius$^1$, Dieter Fox$^{1, 2}$, and Yashraj Narang$^1$
\thanks{The authors are with $^1$NVIDIA Corporation and $^2$University of Washington.\\
$^{\ast}$Equal Contribution}%
}

\maketitle

\begin{abstract}
For both humans and robots, the sense of touch, known as tactile sensing, is critical for performing contact-rich manipulation tasks. Three key challenges in robotic tactile sensing are 1) interpreting sensor signals, 2) generating sensor signals in novel scenarios, and 3) learning sensor-based policies. For visuotactile sensors, interpretation has been facilitated by their close relationship with vision sensors (e.g., RGB cameras). However, generation is still difficult, as visuotactile sensors typically involve contact, deformation, illumination, and imaging, all of which are expensive to simulate; in turn, policy learning has been challenging, as simulation cannot be leveraged for large-scale data collection. We present \textbf{TacSL} (\textit{taxel}), a library for GPU-based visuotactile sensor simulation and learning. \textbf{TacSL} can be used to simulate visuotactile images and extract contact-force distributions over $200\times$ faster than the prior state-of-the-art, all within the widely-used Isaac Simulator. Furthermore, \textbf{TacSL} provides a learning toolkit containing multiple sensor models, contact-intensive training environments, and online/offline algorithms that can facilitate policy learning for sim-to-real applications. On the algorithmic side, we introduce a novel online reinforcement-learning algorithm called asymmetric actor-critic distillation (\sysName), designed to effectively and efficiently learn tactile-based policies in simulation that can transfer to the real world. Finally, we demonstrate the utility of our library and algorithms by evaluating the benefits of distillation and multimodal sensing for contact-rich manipulation tasks, and most critically, performing sim-to-real transfer. Supplementary videos and results are at \url{https://iakinola23.github.io/tacsl/}.
\end{abstract}


\section{Introduction}
\IEEEPARstart{F}{or} humans, the sense of touch is an essential means of perception. Touch sensors cover the surface of the human body and are critical for diverse tasks, such as object recognition, grasping, manipulation, and locomotion \cite{johansson2009coding, rogers2001passive, ryan2021interaction}. In robotics, research has demonstrated that tactile sensors are invaluable when performing analogous tasks, particularly contact-rich manipulation tasks \cite{luo2017robotic, li2020review,  lepora2021soft, she2021cable, zhao2023skill}. Nevertheless, tactile sensors are far less widespread than other sensing modalities (e.g., RGB-D cameras, force/torque sensing), in part due to fundamental difficulties presented by tactile data streams.

In particular, three longstanding challenges are \textit{interpretation} (i.e., mapping from sensor signals to quantities of interest, such as forces and torques), \textit{generation} (i.e., generating sensor signals in novel scenarios), and \textit{policy learning} (i.e., mapping from sensor signals to useful robot actions). For \textit{visuo}tactile sensors \cite{yuan2017gelsight, ward2018tactip, alspach2019soft, lambeta2020digit}, which contain embedded cameras, interpretation has been greatly facilitated by their close relationship with standard vision sensors (e.g., RGB cameras). Thus, researchers have been able to leverage well-established computer vision and image processing algorithms. Nevertheless, generation remains a challenge, as these sensors typically involve contact, large deformations (e.g., the indentation of soft membranes), illumination (e.g., by multiple LEDs), and imaging. Each of these phenomena alone can be expensive to simulate. In turn, policy learning based on tactile sensors has faced bottlenecks, as simulation cannot be efficiently leveraged for large-scale data collection or experience generation.

\begin{figure}[t]
\begin{center}
\includegraphics[width=1.\linewidth]{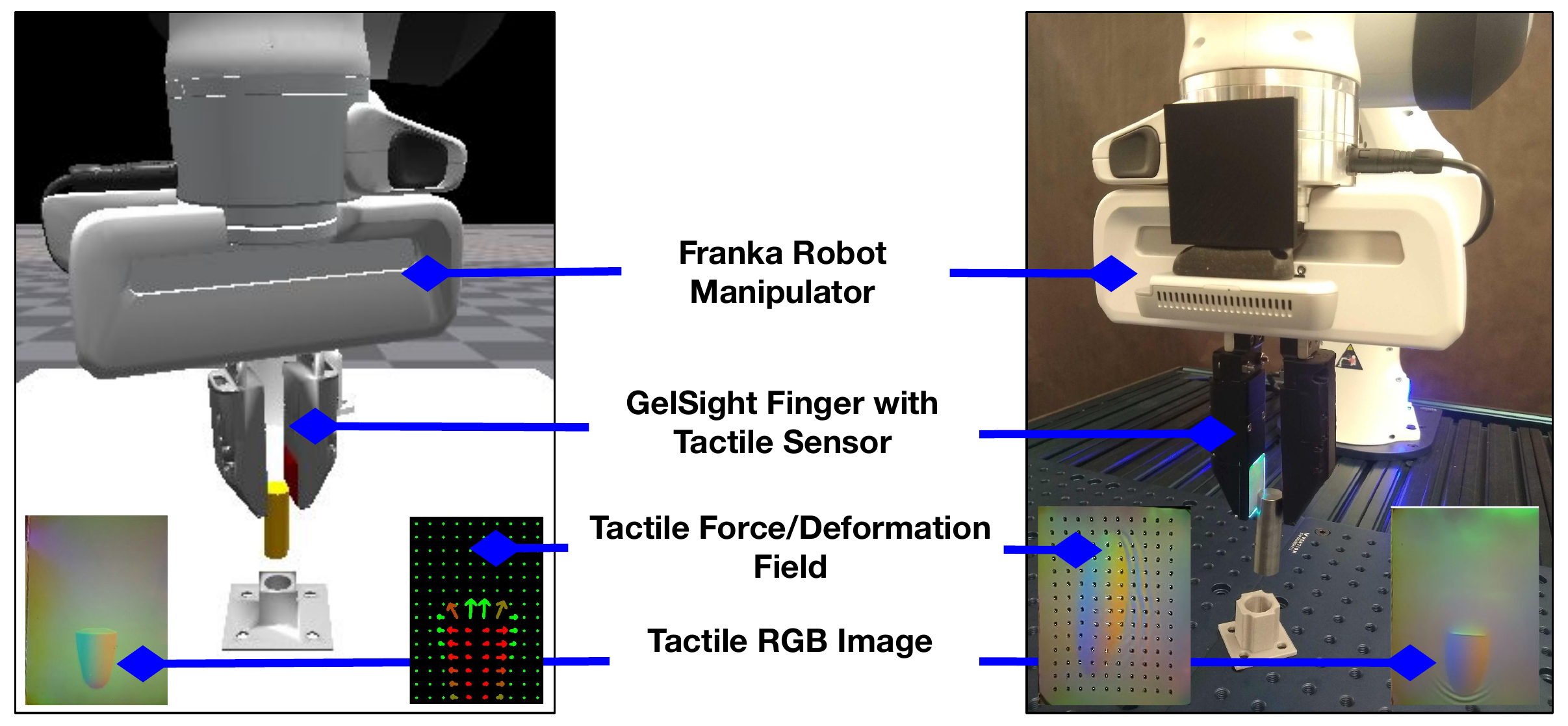}
\end{center}
\caption{
Using state-of-the-art tactile simulation methods, \textbf{TacSL} equips a simulated robot (left) with tactile-sensing capabilities that mirror those available on a real-world robot (right).
By employing algorithms provided within the \textbf{TacSL} learning toolkit, tactile-based policies for contact-rich tasks (e.g., peg insertion) are trained within simulation, thus enabling scalable data collection and preserving the lifespan of the real-world tactile sensor. Subsequently, learned policies can be transferred successfully to the real-world system.
}
\label{fig:tactile_highlight}  
\end{figure}

Several general-purpose robotics simulators have been developed, enabling efficient development and testing of algorithms in perception and control, as well as training of reinforcement learning (RL) policies \cite{makoviychuk2021isaac, todorov2012mujoco, coumans2021, xu2023efficient}. Adjacent to these simulators, specialized tactile simulation modules have also been built, enabling development of touch-based algorithms before incurring the cost of tactile hardware \cite{narang2021sim, si2022taxim, wang2022tacto, lin2022tactile}. Nevertheless, there is a dearth of fast, highly-parallelized visuotactile simulation modules that are integrated with general-purpose simulators; furthermore, tactile-based learning methods that can leverage these simulators have been limited, particularly data-hungry methods involving on-policy RL or online policy distillation.

\begin{figure*}[t]
\includegraphics[width=1.0\textwidth]{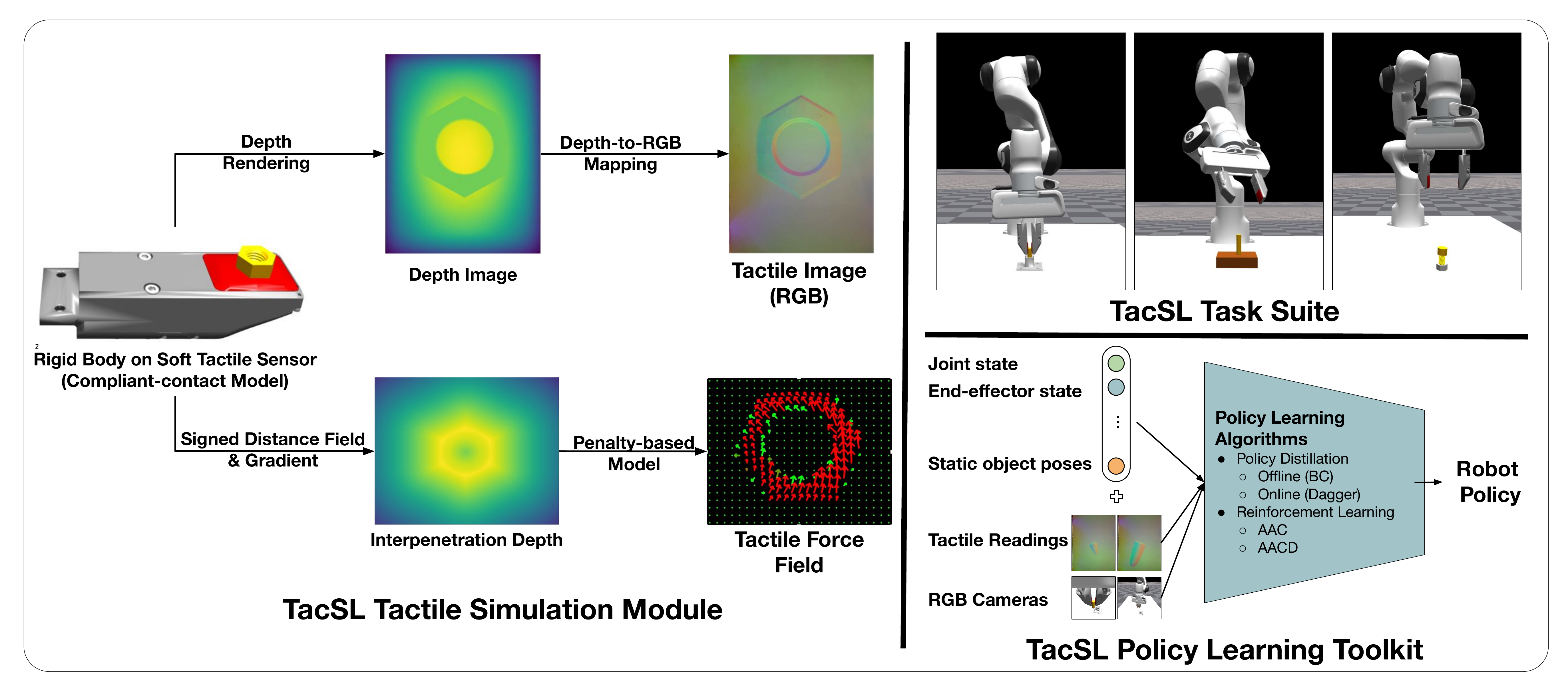}
\caption{\textbf{TacSL} has 3 main components: (Left) A fast visuotactile simulation module that produces  tactile images and force fields. (Top-right) A set of sensors and 
manipulation environments for tactile-based policy learning. (Bottom-right) Offline and online distillation as well as reinforcement learning algorithms to facilitate sim-to-real transfer.}
\label{fig:tacsl_systems}
\end{figure*}

In this work, we present \textbf{TacSL} (pronounced \textit{taxel}) (Figure~\ref{fig:tactile_highlight}), a GPU-based tactile simulation module for visuotactile sensors that is integrated into a general-purpose robotics simulator. We also provide several tools that can facilitate efficient progress in tactile-based policy learning. Finally, we use our simulator and tools to demonstrate sim-to-real transfer. Our specific contributions are the following (Figure~ \ref{fig:tacsl_systems}):

\begin{itemize}%
    \item \textbf{A fast, highly-parallelizable simulation module for visuotactile sensors}. Unlike previous works, the module is fully integrated into a widely-used general-purpose robotics simulator, Isaac Simulator \cite{makoviychuk2021isaac}, and provides both simulated RGB images and normal and shear force fields.
    
    \item \textbf{A set of sensors, environments, and algorithms} to jump-start prototyping and training of tactile-based learning algorithms. These include two models of the GelSight sensor \cite{yuan2017gelsight}, three robotic assembly environments with randomized positions and orientations, and two distillation algorithms for teacher-student policy learning \cite{hwangbo2019learning, chen2022system}, including an online distillation algorithm.
    
    \item \textbf{A novel policy-learning algorithm ({\sysName})} that leverages a pretrained critic to accelerate learning contact-rich policies with high-dimensional inputs such as tactile images. The sample efficiency of {\sysName} enables on-policy RL, even with image augmentation applied to the high-dimensional input.
    
    \item \textbf{An analysis on the utility of tactile sensing for contact-rich manipulation}. The analysis uses the aforementioned simulation module and learning tools and addresses an open question in the research community: how does policy-learning performance differ when training from privileged state information, versus vision, tactile images, and/or multimodal inputs?
    
    \item \textbf{A recipe for sim-to-real transfer.}
    We show how to use the described techniques, including soft-contact parameter randomization and image augmentation during policy learning to train visuotactile policies in simulation that transfer in zero-shot to the real world.
\end{itemize}

We plan to open-source TacSL's tactile simulation module and learning tools. Our aim is to encourage widespread research on tactile-based algorithm development for perception, control, and policy learning, as well as to further increase the adoption of touch sensing across the manipulation community.

\section{Related Works}
\label{sec:citations}

\subsection{Tactile Simulation}

To simulate tactile sensors, two fundamental processes must be simulated: 1) contact interaction between the tactile sensor and an object, which generates physical outputs (e.g., forces, deformation), 2) transduction of these physical outputs to the actual tactile measurements (e.g., electrical signals \cite{wettels2008biomimetic}, magnetoelectric signals \cite{bhirangi2021reskin}, or visuotactile images \cite{johnson2009retrographic}). 

For visuotactile sensors, process 1 (contact interaction) involves the indentation and deformation of elastomeric membranes. Gold-standard simulation of elastomeric deformation is typically achieved via the finite element method (FEM) \cite{Reddy2019Book}. However, scientific FEM simulators can take exceptionally long to execute \cite{narang2021interpreting, narang2021sim}, less-accurate robotics implementations still execute slower than real-time \cite{ma2019dense, sferrazza2019ground, sferrazza2020learning, narang2021sim, huang2022defgraspsim}, and faster neural approximations require a large amount of domain-relevant training data \cite{pfaff2020learning, huang2023defgraspnets}. 
A recent work \cite{10388459} developed a FEM-based physics engine for simulating the tactile sensor deformation accurately and efficiently. However, the authors identified several limitations, including low simulation speed when handling complex rigid bodies and the lack of integration with a general-purpose simulator.

To accelerate simulation, recent efforts have modeled these deformations with rigid-body approximations. 
This process consists of two steps: 1) rigid-body contact is simulated between the elastomer and a rigid object to obtain plausible interaction forces, and 2) the interaction forces are used with a prescribed stiffness coefficient to obtain a plausible interpenetration depth, which approximates the maximum deformation of the elastomer~\cite{wang2022tacto}. As an alternative, a soft contact model can be used, allowing more accurate modeling of statics and dynamics \cite{elandt2019pressure, xu2023efficient}.
In this work, we model contact interaction with a soft contact model in which objects are still modeled as rigid bodies, but can interpenetrate in proportion to interaction forces. The interpenetrated state is used as input to the simulated imaging process, while both the interpenetrated state and relative body velocities are used to compute contact force fields. In contrast to prior work \cite{xu2023efficient}, our physics simulation module is GPU-accelerated, achieving speed-ups of $300\times$.

For visuotactile sensors, process 2 (transducing deformations to visuotactile images) involves direct RGB imaging of the deformed surface of the elastomer. Some simulators have simplified this problem by directly rendering depth images of the object \cite{church2022tactile}. Other efforts have simulated RGB images through a classical illumination model \cite{gomes2021generation}, a calibrated polynomial look-up table \cite{si2022taxim, wang2022tacto}, or learned generative models, including GANs \cite{jing2023unsupervised,chen2022bidirectional} and diffusion models \cite{higuera2023learning}. In contrast to prior works \cite{si2022taxim, wang2022tacto}, our image rendering is integrated into our parallelized physics simulator without any I/O to an external renderer, achieving speed-ups of over $200\times$.

\subsection{Tactile Policy Learning}

Tactile simulators have been used for a diverse set of tasks, such as perceptual tasks (e.g., object identification, contact location prediction, and slip detection \cite{ding2020sim, narang2021sim}), non-prehensile manipulation tasks (e.g., pushing, edge following, and rolling)\cite{church2022tactile}, prehensile manipulation tasks (e.g.,  grasping \cite{wang2022tacto,do2023densetact} and insertion \cite{xu2023efficient}), and bimanual tasks~\cite{lin2023bi}.

Within grasping and manipulation, a number of works have aimed to learn tactile-based \textit{policies} that can map tactile observations (e.g., visuotactile images) to robot actions (e.g., joint torques or pose targets). For visuotactile sensors, the majority of simulators have been utilized for supervised learning, where datasets are first generated and learning algorithms are subsequently applied to the dataset.
Fewer works have explored using online learning algorithms (e.g., online RL methods) for policy learning, which typically require fast and efficient data generation \cite{bi2021zero, church2022tactile, xu2023efficient}. Among these, \cite{bi2021zero} 
demonstrated sim-to-real transfer on swing-up manipulation; \cite{xu2023efficient} demonstrated sim-to-real transfer on a peg insertion task; and \cite{church2022tactile} learned depth-based policies, converting RGB to depth during real-world deployment, and performed sim-to-real transfer on non-prehensile tasks.

In this work, we use our simulator and renderer for model-free, on-policy RL, as well as offline and online policy distillation. In contrast with \cite{bi2021zero}, we enable policy training for widely-available visuotactile sensors rather than custom models; in comparison with \cite{xu2023efficient}, we implement multiple policy-learning algorithms and sensors; and in contrast with \cite{church2022tactile,lin2022tactile}, we extract normal and shear force distributions, simulate realistic RGB images, and demonstrate prehensile manipulation.

Concurrent work \cite{qi2023general} has explored combining tactile images and third-person camera views for dexterous manipulation. The work obtains low-dimensional contact location from the tactile image. In contrast, our work focuses on greatly accelerating tactile simulation needed for end-to-end tactile policy learning, providing a comprehensive policy learning toolkit with different algorithms, and describing a recipe for sim-to-real transfer of end-to-end policies that consume raw tactile images. \textbf{TacSL} can thus be a valuable testbed to accelerate developing and evaluating new tactile algorithms for efforts such as \cite{qi2023general}. 

To the best of our knowledge, \textbf{TacSL} is the first general-purpose simulation module that provides fast and efficient simulation of both tactile image and tactile force-field sensing. It achieves the necessary speed for online learning of \textit{end-to-end} tactile-image-based policies for prehensile contact-rich manipulation, distinguishing it from previous works. \textbf{TacSL} also provide the algorithmic tools enabling others to perform effective policy learning and sim-to-real transfer.

\section{Fast Visuotactile Simulation}
\label{sec:tactile_simulation}

\textbf{TacSL} simulates visuotactile sensors in two phases. First, it simulates the physical interactions between the tactile sensor and indenting objects in a fast and stable manner. Based on the simulation, \textbf{TacSL} then extracts and computes two tactile measurements: tactile RGB images and tactile force fields. Notably, both phases leverage GPU parallelization to achieve substantial performance improvements compared to existing state-of-the-art approaches. We now describe these simulation components of \textbf{TacSL} in detail. 

\subsection{Contact Simulation}
\label{sec:contact_simulation}
This section describes how dynamic contact effects are handled in \textbf{TacSL}. We 1) outline our general dynamics solver procedure, 2) explain the governing analytical equations of our soft contact constraints, and 3) address how these equations are solved in a numerically-stable manner.
The contact constraint equations derived here are implemented within the PhysX SDK~\cite{PhysX} physics engine, whose built-in solver procedure (outlined below) ensures constraint-consistent dynamics. Collision geometry and contact parameters are configured on the \textbf{TacSL} side.

\textbf{Dynamics Solver:} The PhysX dynamics solver operates on a discrete-time formulation with a discretization interval~$\Delta t$.
Given the position $p$ and velocity $v$ of a body at the current time-step, the solver computes the position $p^+$ and velocity $v^+$ at the next time step using a semi-implicit Euler integration scheme:
\begin{align}
v^+ &\xleftarrow{}  v + \Delta v \label{eq:semi_implicit_euler_v} \; , \\
p^+ &\xleftarrow{}  p + \Delta t \, v^+ \label{eq:semi_implicit_euler_p} \; ,
\end{align}
where the velocity change $\Delta v$ is the combined effect of external and constraint forces on that body.

We use a Gauss-Seidel-style solver due to its simplicity and fast convergence, based on~\cite{macklin2019small}. The solver subdivides the frame duration into a configurable number of smaller timesteps and applies a sequential impulse strategy for each of these substeps.
Each constraint (e.g., a collision) computes an impulse to reduce the constraint error and applies it to the system by changing $\Delta v$ of the involved bodies.
The procedure for one time frame is given in Algorithm~\ref{alg:TGS_solver}, with further details of the dynamics solver provided in Appendix~\ref{sec:appendix_dyn_solver}.

\begin{algorithm}
\caption{Dynamics Solver Step (Gauss-Seidel)}\label{alg:TGS_solver}
\begin{algorithmic}[1]
\Require Initial body positions and velocities
\For{substep $n=0,1,...,N-1$}
    \For{constraint $j=0,1,...,C-1$}
        \State Compute the impulse $\lambda$ that is to be applied over this substep based on the most up-to-date estimate of the positions and velocities, ignoring any other constraints.
        \State Update the body velocities based on the computed impulse.
    \EndFor
    \State Integrate body positions based on the updated velocities over the timestep $\Delta t/N$.
\EndFor
\end{algorithmic}
\end{algorithm}

\textbf{Contact Constraints:}
The one constraint type of interest here is soft contact constraints
to simulate the interaction between the deformable elastomeric membrane of a tactile sensor and an indenting object.
The membrane and object are both modeled as rigid bodies, but strict non-penetration constraints are replaced with penalty-based constraints to recover the softness.
Specifically, we use a Kelvin-Voigt model, which consists of a spring and a damper connected in parallel \cite{fliigge1967viscoelasticity,kao2016contact}. 
The unilateral contact force $f$ at each contact point is given by: 
\begin{equation}
\label{eq:kelvin_voigt}
f =  \max(-\kappa \epsilon - c \dot{\epsilon}, 0) \; ,
\end{equation}
where $\kappa$ and $c$ are stiffness and damping constants, respectively, $\epsilon$ is the contact distance between membrane and object, and $\dot{\epsilon}$ is the separation velocity.

The number of contact points between sensor and object is dynamically computed based on the collision geometry.
Each contact point acts independently in normal direction, which is presented in the following paragraph.
Contact points with similar normal direction are grouped together in a contact patch for the purpose of computing a single Coulomb friction force.

\begin{figure}[h]
\begin{center}
\includegraphics[width=1.\linewidth]{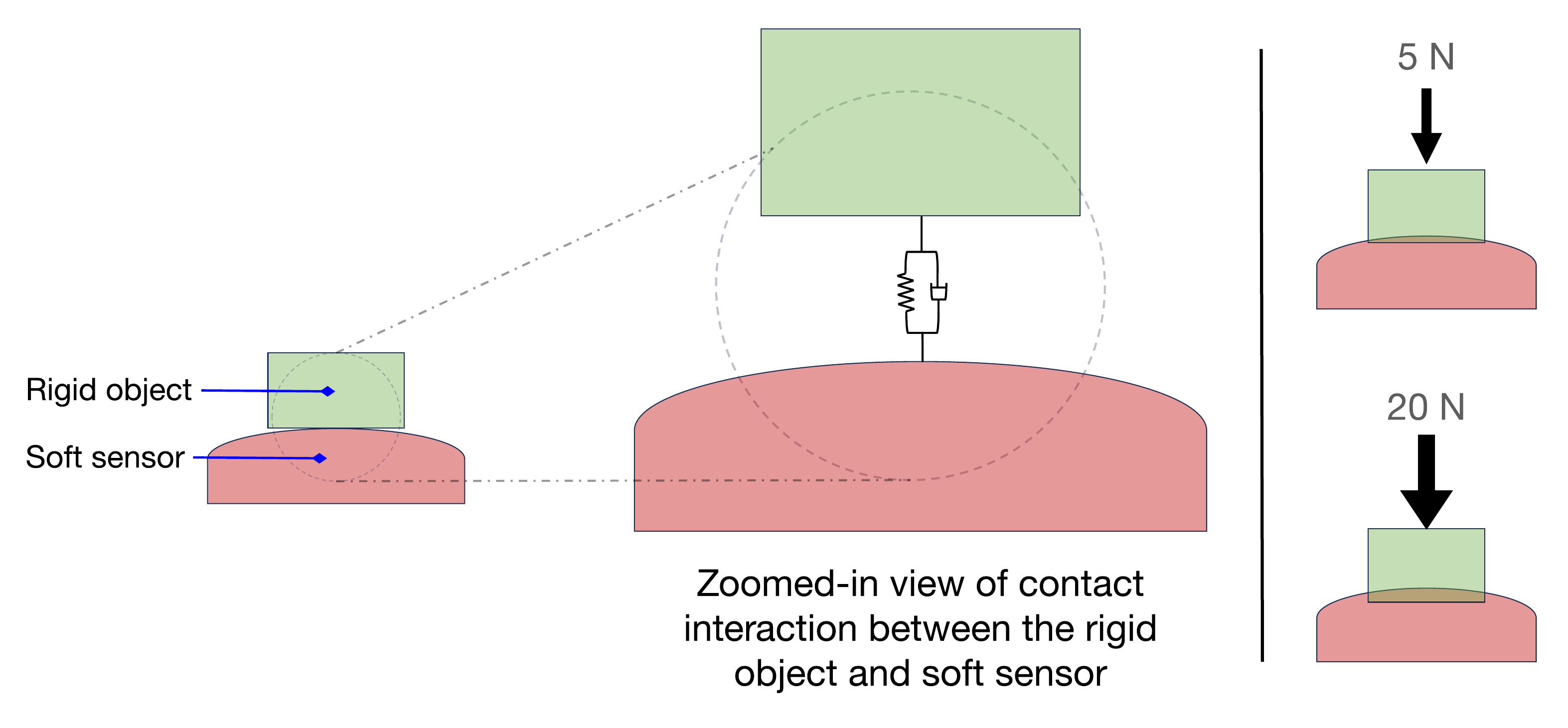}
\end{center}
\caption{
Illustration of the soft contact model. Objects are modeled as rigid bodies, interpenetration constraints  are relaxed for the soft object, and a level of interpenetration is allowed according to a spring-damper system. Right: the level of interpenetration scales with magnitude of the applied force.
}
\label{fig:compliant_contact}  
\end{figure}

\textbf{Constraint Solution:} In an exemplary 1D scenario, the discrete contact dynamics are given as
\begin{equation}
\lambda =  m \Delta\dot{\epsilon} \; , \label{eq:discrete_contact_dynamics}
\end{equation}
where $\lambda = f \Delta t$  is the contact impulse, $\Delta\dot{\epsilon}$ is the velocity change over the time step, and $m$ is the effective inertia at the contact point.

Using Eqns.~\ref{eq:kelvin_voigt} and~\ref{eq:discrete_contact_dynamics} directly to compute velocity updates in Alg.~\ref{alg:TGS_solver}
would be notoriously unstable for stiff springs.
Therefore, we numerically handle the soft contact as an implicit spring\cite{tan2011stable}, which enhances stability and is mathematically equivalent to an implicit Euler step for this constraint.
To this end, the discretized spring equation must be fulfilled \textit{at the end} of a time step, i.e.,
\begin{equation}
\lambda = \Delta t \max(-\kappa \epsilon^+ - c \dot{\epsilon}^+, 0) \; . \label{eq:implicit_impulse}
\end{equation}

From the semi-implicit Euler integration scheme~(Eqns.~\ref{eq:semi_implicit_euler_v} and~\ref{eq:semi_implicit_euler_p}), we have
\begin{align}
    \dot{\epsilon}^+ &= \dot{\epsilon} + \Delta \dot{\epsilon} \; , \\
    \epsilon^+ &= \epsilon + \Delta t \dot{\epsilon}^+ = \epsilon + \Delta t (\dot{\epsilon} + \Delta \dot{\epsilon}) \; .
\end{align}

Combining these equations with the contact dynamics (Eqn.~\ref{eq:discrete_contact_dynamics}) and the force law (Eqn.~\ref{eq:implicit_impulse}), we obtain a solvable (causal) expression for the contact impulse~$\lambda$.
We omit the $\max$ operator for brevity:
\begin{align}
\lambda &=  -\Delta t(\kappa (\epsilon + \Delta t(\dot{\epsilon} + \Delta \dot{\epsilon})) + c (\dot{\epsilon} + \Delta \dot{\epsilon})) \nonumber \\
&=  -\Delta t \kappa \epsilon - \Delta t(\Delta t \kappa + c ) (\dot{\epsilon} + \lambda/m)) \; .
\end{align}
The expression incorporates the effect of evaluating the force on the end-of-timestep contact position and velocity, thereby avoiding overshoot.

Solving for $\lambda$, we obtain:
\begin{align}
\label{eq:contact_impulse_solve}
\lambda = \dfrac{-\Delta t \kappa \epsilon - \alpha \dot{\epsilon}}{1 + \alpha /m} \; ,
\end{align}
where $\alpha :=  \Delta t(\Delta t \kappa + c )$.
This last equation~(Eqn.~\ref{eq:contact_impulse_solve}) contains only quantities known at the beginning of the time step and is suitable for implementation within Alg.~\ref{alg:TGS_solver}.

\subsection{Tactile Image Generation}

As opposed to traditional tactile sensors, visuotactile sensors return RGB images. However, in simulation, it can be prohibitively challenging to tune the light sources within the sensor (e.g., multi-colored LEDs) and optical properties of the sensor (e.g., a semi-transparent membrane) in order to render a realistic non-uniform light distribution. Thus, we instead render depth images in the simulator and map the depth image to an RGB image. Specifically, following prior approaches \cite{wang2022tacto, si2022taxim}, we place a simulated camera at the effective optical location inside the tactile sensor and render a depth image $I_{depth}$ showing the indenting object. We then map $I_{depth}$ to an RGB image $I_{rgb}$ via a relationship $I_{rgb} = F(I_{depth})$, where $F$ is parameterized by a calibrated look-up table \cite{si2022taxim}.
Importantly, unlike previous works (e.g., \cite{wang2022tacto}), the rendering occurs in parallel within the simulator without any I/O to an external renderer.

\subsection{Normal/Shear Force-Field Computation}
\label{sec:force_field_simulation}
Visuotactile sensors not only provide geometric information about the object, but also contain information about normal and shear force distributions during contact \cite{sferrazza2019ground, ma2019dense}. These distributions describe pressure, shear, torsion, and slip along the tactile membrane. To compute these force distributions, we adopt the penalty-based tactile model from \cite{xu2023efficient}.
Tactile points are sampled across the surface of the sensor (see example in Fig.~\ref{fig:tactile_examples}), and the contact force at each tactile point is computed as:
\[
\mathbf{f}_n = (-k_n + k_d \dot{d})d\mathbf{n}, \quad \mathbf{f}_t = -\frac{\mathbf{v}_{t}}{||\mathbf{v}_{t}||} \min (k_t||\mathbf{v}_{t}||, \mu ||\mathbf{f}_n||)
\]
where $\mathbf{f}_n$ is the contact normal force; $k_n$ and $k_d$ are the contact stiffness and contact damping; $d$ and $\dot{d}$ are the interpenetration distance and velocity; 
$\mathbf{n}$ is the contact normal; $\mathbf{f}_t$ is the frictional force; $\mathbf{v}_t$ is the tangential velocity; and $k_t, \mu$ are the friction stiffness and coefficient of friction, respectively.\footnote{Note that the inclusion of $d$ in the contact normal damping force follows the formulation in \cite{xu2021end}; please see the paper for further details.}

Prior to simulation, we compute the signed distance field (SDF) of the contacting object as described in \cite{narang2022factory, macklin2020local}. At each timestep, the interpenetration distance $d$ is obtained by querying the SDF of the contacting object at the tactile points. To get the interpenetration velocity $\dot{d}$, we use chain rule to rewrite $\dot{d} = (\mathbf{\nabla}{d})^T \mathbf{\dot{x}}$, where $\mathbf{\dot{x}}$ is the relative velocity at the contact point. The contact normal $\mathbf{n}$ is
defined as the gradient of the SDF values: $\mathbf{\nabla}{d} = \mathbf{n}$. 
The SDF gradient is calculated using finite differencing. Notably, both SDF queries and gradient computations are computationally efficient and easily parallelizable.

Previous efforts (namely, \cite{xu2023efficient}) have computed these normal and shear forces serially on CPU; to improve simulation speed, we instead compute the forces in parallel on GPU. The gains from our parallelized approach scale with the number of tactile locations on the sensor (i.e., the density of queried locations on the tactile surface) for which contact forces are computed, as well as the number of sensors simulated in a multi-environment learning setup. 
In addition, \cite{xu2023efficient} only handles objects represented by primitive shapes (e.g., cuboids, cylinders, etc.), whereas our simulation module can handle objects represented by arbitrary meshes.

\section{Efficient Visuotactile Policy Learning}
\label{sec:tactile_learning}
To jump-start prototyping and training of tactile-based learning algorithms, we provide implementations of different approaches for policy learning, which can be used for contact-rich tasks such as peg insertion and nut-and-bolt fastening.
These approaches are facilitated by our parallelized simulation module, which enables fast acquisition of experience required for on-policy learning algorithms. 
All the presented algorithms rely on a pretrained RL expert that has low-dimensional inputs to the actor and critic.

Specifically, we provide two families of learning algorithms. The first family of algorithms is \textit{policy distillation}, which trains a student policy to mimic examples provided by a pretrained expert. We provide both offline and online variants of policy distillation. The second family of algorithms is \textit{rewards-based learning} which employs RL to train policies with high-dimensional inputs. Here, we provide the asymmetric actor-critic approach to high-dimensional RL. In addition, we introduce a novel RL algorithm (\sysName) that leverages a pretrained critic of a low-dimensional state-based RL agent to efficiently train a high-dimensional image-based RL agent.

\textbf{Preliminaries}: We model the contact-rich manipulation task as a Markov decision process (MDP) defined by $(\mathcal{S}, \rho_0, \mathcal{A}, R, \mathcal{T}, \gamma)$ for the state space $\mathcal{S}$, initial state distribution $\rho_0$, action space $\mathcal{A}$, reward function $R(s, a, s')$, transition distribution $\mathcal{T}(s', s, a) $, and discount factor $\gamma$. Importantly, we distinguish between the state $s \in \mathcal{S}$, which contains the full state of the robot and environment (e.g., object poses, contact forces), and the measurable observation $o \in \mathcal{O}$, which contains measurements accessible via real-world sensors (e.g., camera images, tactile images). 
The objective of policy learning is to obtain a policy $\pi: a = \pi(o)$ that maps from observation to action in order to maximize the expected sum of rewards.

\subsection{Policy Distillation}
\label{sec:policy_distillation}

Policy distillation employs a \textit{teacher-student} framework for policy learning where the behavior of a trained policy network (the teacher) is transferred to a different policy model (the student)\cite{torabi2018behavioral,ross2011reduction,rusu2015policy,czarnecki2019distilling}. %
Here, we first leverage an RL algorithm (PPO \cite{schulman2017proximal}) to learn high-performing state-based \textit{teacher} policies $\pi_{e}: a = \pi_{e}(s)$ for contact-rich tasks; these policies take as input privileged information that is available only in simulation, including exact rigid body poses and  pairwise net contact forces between bodies in the scene. Then, we distill these state-based policies into  \textit{student} policies $\pi_{s}: a = \pi_{s}(o)$; these only take as input tactile images and proprioceptive information that are available in the real world.

\begin{figure*}
\centering
  \includegraphics[width=1.0\textwidth]{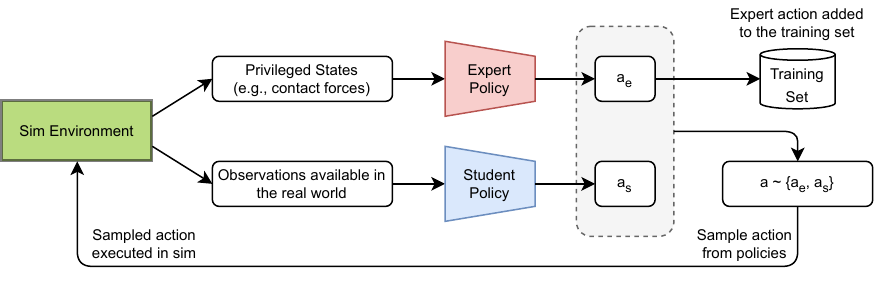}
  \caption{\textbf{Tactile Policy Distillation} is a method to efficiently train a policy that takes as input high-dimensional visuotactile observations. %
  An expert policy $\pi_{e}$ is first trained to solve the task; this policy takes as input  privileged state information available only in the simulation (e.g., contact forces) and predicts action $a_e$.  During distillation, a student policy is trained to imitate the expert policy; this policy takes as input observations that are available in both simulation and the real world (e.g., visuotactile images) and predicts action $a_s$. The expert action $a_e$ is always used as training data for the student action $a_s$. Nevertheless, to advance the simulator to the next step, an action is sampled from either the expert policy or the student policy.}
\label{fig:distillation_illustration}
\end{figure*}

Two standard approaches to policy distillation are implemented in \textbf{TacSL}.
First, offline distillation, also referred to as behavior cloning (BC), learns to mimic actions from a fixed dataset of simulated experience. The simplicity of BC enables computationally-cheap and fast implementations; however, its performance is limited to the size and quality of the offline dataset. 
Second, \textbf{TacSL} provides an online distillation approach, also known as DAgger\cite{ross2011reduction}, where the student policy is increasingly used to collect more experience as training proceeds. For completeness, we describe our application of the approach in detail in Algorithm~\ref{alg:cap} and Fig.~\ref{fig:distillation_illustration}.

Specifically, the expert policy $\pi_{e}$ is used to collect trajectories with probability $\beta$, while actions are sampled from the imperfect student policy $\pi_s$ with probability (1 - $\beta$). As learning proceeds, $\beta$ decreases over time.
During this process, for each visited state, the actions suggested by the expert policy are recorded as the correct label. Note that keeping $\beta$ constant at $1$ reduces the algorithm to the static case of behavior-cloning the expert in an iterative fashion with a continuously updated dataset.

\begin{algorithm}
\caption{Tactile Policy Distillation}\label{alg:cap}
\begin{algorithmic}[1]
\Require Expert policy $\pi_{e}$
\State Initialize student policy $\pi_{s}$, fraction of time the agent uses the action from expert policy $\beta = 1$ 
\For{iteration $n=0,1,...,N-1$}
    \State Update $\beta$ according to schedule \Comment{$\beta$ reduces over time}
    \For{iteration $m=0,1,...,M-1$}   \Comment{Collect $M$ rollout trajectories \{$\tau_0, ..., \tau_{M-1}$\}, given $\pi_{e}$, $\pi_{s}$ and  $\beta$}
        \State $\tau_m = \{\}$ 
        \For{timestep $t=0,1,...,T-1$}
            \State $a_s = \pi_{s} (o)$ 
            \State $a_e = \pi_{e} (s)$        
            \State $a = $ choose from $ {a_s, a_e}$ according to probability $\beta$     
            \State Execute action in environment
            \State $\tau_m = \tau_m \cup \{(s, a_e)\}$  \Comment{Always record expert label}
        \EndFor
    \EndFor
    \For{iteration $e=0,1,...,E$}
        \State Train  $\pi_{s}$ with rollout trajectories in dataset buffer
    \EndFor
    \State Evaluate student policy $\pi_{s}$ \Comment{Obtain success rate}
\EndFor
\end{algorithmic}
\end{algorithm}

\subsection{Reinforcement Learning}

Whereas policy distillation learns from supervision labels provided by an expert, reinforcement learning learns from reward feedback obtained via environment interactions.
Enabled by our fast tactile simulation module, we also show that we can learn skills for contact-rich manipulation tasks via on-policy RL. For this purpose, we use proximal policy optimization (PPO) \cite{schulman2017proximal} to solve different contact-rich tasks and compare performance over different sensing modalities.
 
\subsubsection{Asymmetric Actor-Critic}
To learn policies with high-dimensional sensor inputs available in a real-world setting, we employ an asymmetric actor-critic framework \cite{pinto2017asymmetric}. In this setup, the critic leverages privileged information available in simulation, such as the accurate poses of all robot and object parts, as well as distributed contact forces. Conversely, the actor operates with tactile images and observation modalities that are accessible in the real world, such as robot joint angles. 
This approach improves computational and sample efficiency, and as a result scalability of reinforcement learning (RL) to high-dimensional state or observation spaces. As the critic is a much smaller network with lower-dimensional inputs, it trains faster and offers better estimates of the Q-values for training the larger policy network during the policy improvement step. 
Once trained, the actor can be deployed on the real robot, as its input is readily available in the real world.

\subsubsection{Asymmetric Actor-Critic Distillation (\sysName)}
\label{sec:aacd}
While contact-rich tasks such as insertion require force to accomplish the task, it is also important to minimize excessive contact forces during task execution to prevent damage to both the parts and the robot. As a result, minimizing excessive contact forces while solving contact-rich tasks poses a challenging trade-off for exploration, especially with high-dimensional inputs such as tactile images.

\begin{algorithm}
\caption{Asymmetric Actor-Critic Distillation}\label{alg:aacd}
\begin{algorithmic}[1]
\Require initial low-dim policy parameters $\theta_{s}^{0}$, low-dim value function parameters $\phi_{s}^{0}$, and high-dim policy parameters $\theta_{o}^{0}$. All parameters are randomly initialized.

\State \textbf{Stage 1:} Train low-dim policy $\pi_{\theta_s}(s)$ and value function $V_{\phi_s}(s)$
\For{$n=0$ to $N-1$}
    \State Collect rollout data with $\pi_{\theta_{s}^{n}}(s)$
    \State Update $\theta_{s}^{n+1}$ using PPO objective and $V_{\phi_{s}^{n}}(s)$
    \State Update $\phi_{s}^{n+1}$ using Bellman loss
\EndFor

\State \textbf{Stage 2:} Train high-dim policy $\pi_{\theta_o}(o)$ and fine-tune $V_{\phi_s}(s)$
\For{$m=0$ to $M-1$}
    \State Collect rollout data with $\pi_{\theta_{o}^{m}}(o)$
    \State Update $\theta_{o}^{m+1}$ using PPO objective and $V_{\phi_{s}^{N+m}}(s)$
    \State Update $\phi_{s}^{N+m+1}$ using Bellman loss
\EndFor

\State Evaluate $\pi_{\theta_o^M}$ \Comment{Obtain success rate}
\end{algorithmic}
\end{algorithm}

\begin{figure}%
\begin{center}
\includegraphics[width=0.95\linewidth]{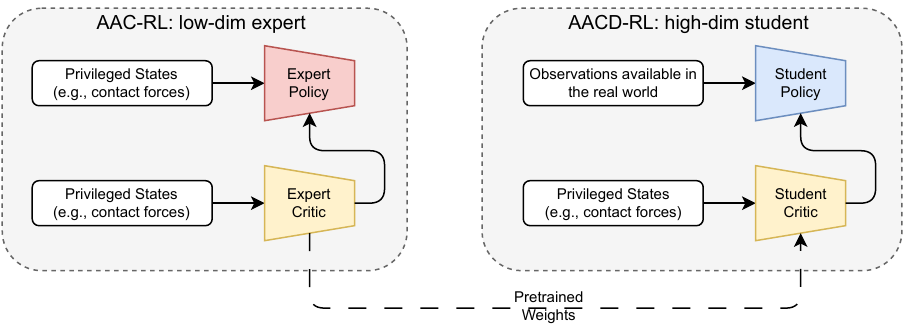}
\end{center}
\caption{\textbf{Asymmetric Actor Critic Distillation (\sysName)}: Illustration of the two stages of \sysName. In the first stage, an expert agent (actor and critic) is trained using RL to learn the task using privileged information available in simulation. In the second stage, the critic is initialized with the pretrained ``expert'' critic. The high-dimensional student policy is similarly trained using RL, as the critic is fine-tuned. 
This approach retains the performance benefits of RL to acquire high dimensional policies with reward-maximizing behaviors.
}
\label{fig:aacd_algo}  
\end{figure}

Inspired by the \textit{teacher-student} framework described earlier, we present a novel asymmetric actor-critic distillation algorithm (AACD) that addresses this exploration challenge by leveraging a pretrained critic from a low-dimensional agent to guide the learning process of a new high-dimensional agent. The low-dimensional agent employs a privileged state-based policy, while the high-dimensional agent operates with image-based observations. %
{\sysName} trains a high-dimensional policy in two steps (see Fig.~\ref{fig:aacd_algo} and Algorithm~\ref{alg:aacd}). First, a randomly-initialized RL actor and critic with  low-dimensional observation input are trained from scratch using low-dimensional privileged state information. Next, a randomly-initialized actor with high-dimensional observation input and the \textit{pretrained} low-dimensional critic are trained and fine-tuned, respectively, to optimize RL policy objectives. {\sysName} effectively addresses exploration challenges in high-dimensional policy learning by leveraging prior knowledge of the pretrained critic, thereby providing effective guidance for learning the high-dimensional policy.

\section{Visuotactile Sim-to-Real Transfer}
\label{sec:bridging_sim2real}

We present key strategies employed to facilitate the successful transfer of tactile policies from simulated environments to the real world. We discuss critical aspects including 1) soft-contact parameter randomization to account for imperfect physics modeling, 2) tactile image augmentation to account for variabilities in tactile camera calibration and other optical variations across different sensors, and 3) the combination of RL plus policy distillation as a strategy to decouple physics parameter randomization during reinforcement learning from image-based randomization during distillation.

\subsection{Physics Parameter Randomization}
To account for the variation in elastomer compliance across sensors and potential changes over time, we randomize the parameters of the soft contact model of our simulator. The stiffness and damping parameters of our compliant contact model  determine the level of softness and velocity decay rate, respectively, of the soft contacts in simulation. These parameters are empirically tuned to reasonable values and are then randomized during policy training (see randomization details in Appendix \ref{sec:training_details}). This randomization ensures that the agent can adapt to uncertainties in the model parameters and variations in the real-world environment.

\subsection{Tactile Image Augmentation}
\label{sec:tactile_image_aug}
\begin{figure}[h]
\begin{center}
\includegraphics[width=0.98\linewidth]{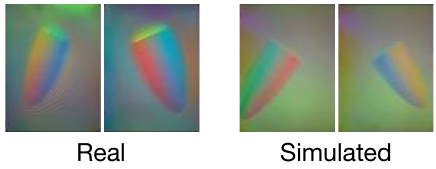}
\end{center}
\caption{
Tactile readings acquired by a robot grasping a peg with a GelSight on each side of a parallel-jaw gripper. Left pair shows real-world images and right pair shows simulated images. Geometry information is very similar across all images, whereas coloration varies between simulation and reality, as well as between two real-world sensors.
}
\label{fig:tactile_image_color_gap}  
\end{figure}

\begin{figure}[h]
\begin{center}
\includegraphics[width=1.0\linewidth]{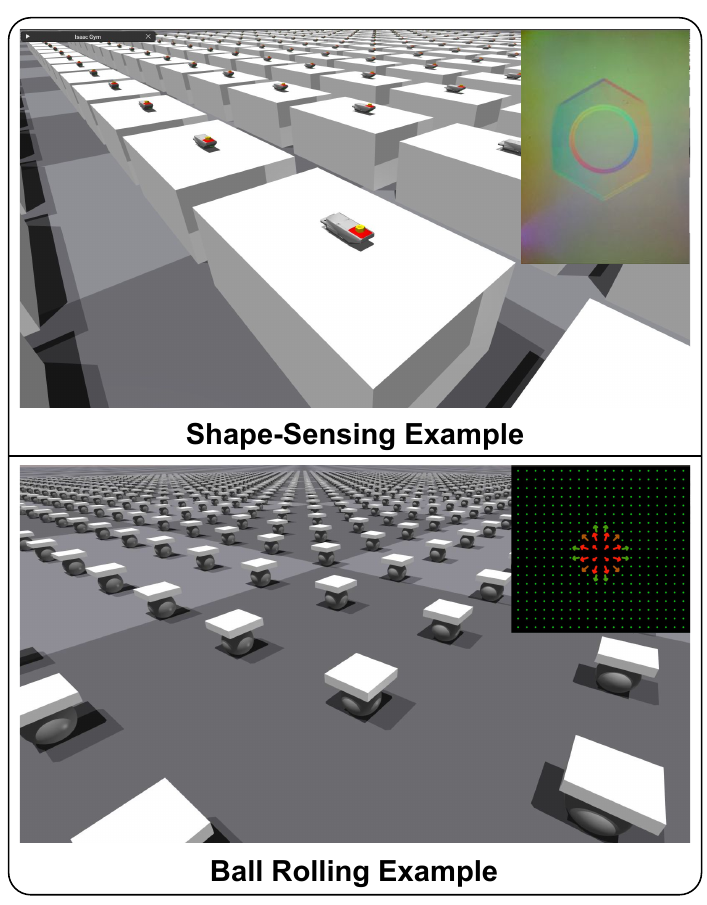}
\end{center}
\caption{\textbf{Shape Sensing (Top)}: An object (hex nut) is pressed against the surface of the GelSight sensor. Inset shows the corresponding visuotactile image. \textbf{Ball Rolling (Bottom)} Adapted from \cite{xu2023efficient}, the surface of the rectangular tactile sensor presses against a sphere and rolls it against the ground plane. Inset shows the extracted tactile force field.}
\label{fig:tactile_examples}  
\end{figure}

In the real world, visuotactile sensors have a physical camera placed behind the elastomer to either directly observe the impression being made on the sensor or indirectly observe it through a mirror. In addition, light sources are usually placed around the sensor to illuminate the viewing area.\footnote{See Figure 3 of Wang et al. \cite{wang2021gelsight} for design details of the GelSight R1.5, a representative visuotactile sensor.} 
In simulation, a camera sensor is placed at the virtual camera location within the simulated environment to mimic the location of the real-world camera. 
Due to manufacturing and assembly variations, the exact camera pose, light locations and lighting intensities can vary between sensors. Illustrated in Fig.~\ref{fig:tactile_image_color_gap}, such variations can be observed between the left and right sensors on a robotic manipulator operating in a real-world environment.  Achieving precise calibration of simulated camera parameters from real-world data can be challenging for a single sensor, let alone for multiple sensors.

As a result, we embrace image augmentation as a scalable and effective way to mitigate imperfections in simulated camera extrinsics and intrinsics, as well as other optical variations inherent to different sensors. 
Specifically, to account for imperfect camera parameters, we apply \textit{spatial randomization} to the tactile images, consisting of random translational shifts and zoom operations, enhancing the model's robustness to variations in sensor camera extrinsics and intrinsics.
In addition, we apply \textit{color randomization}, where we randomize brightness, contrast, saturation, hue, and order of color channels. For each simulated episode, an augmentation transform is sampled and applied throughout the episode, representing the variation introduced when using a new visuotactile sensor in the real world. A reduced level of color augmentation is also applied per timestep.

\subsection{Two-Stage Policy Learning}

The extensive domain randomization needed for sim-to-real transfer, which includes physics randomization and high-dimensional tactile image augmentation, presents increased challenges for policy learning. For example, randomization in high-dimensional spaces amplifies the already-substantial sampling requirements necessary for learning meaningful policies. 
The teacher-student framework employed by the algorithms described in Sections \ref{sec:policy_distillation} and \ref{sec:aacd} makes acquiring transferable policies tractable by improving sample efficiency and enhancing exploration during RL.
In the first stage, a low-dimensional expert policy and critic are trained using physics-based parameter randomization, which leverages privileged low-dimensional information to solve the task efficiently. Subsequently, the expert policy or critic is distilled into a tactile-based policy. 
For this second stage, experience data is generated for policy learning, and the associated image observations are post-processed on-the-fly using the image augmentation techniques described in Section \ref{sec:tactile_image_aug}.

We found the strategies outlined above to be scalable and effective in facilitating the successful transfer of tactile policies from simulation to the real world.

\section{Experimental Results}
\label{sec:result}
Our experiments aim to answer three sets of questions:
\begin{enumerate}[label=\Alph*)]%
    \item How fast is our simulator compared to existing baselines?
    \item Can we learn performant tactile policies in simulation using online distillation algorithms? How does a pretrained critic affect RL with high-dimensional inputs? How does learning from tactile sensing compare to learning from wrist-mounted camera images, as well as multimodal inputs?
    \item Can we effectively transfer policies trained in simulation to a real-world robot?
\end{enumerate}

\subsection{Tactile Simulation Speed}
We measure the the simulation speeds of our simulator for both tactile modalities and compare them to the state-of-the-art tactile simulators for each tactile modality.
\begin{enumerate}%
    \item \textbf{Tactile Image}. We design an object-shape-sensing experiment to evaluate the speed of tactile image generation in our simulator. 
    Here, an object presses against a tactile sensor at a specified location and orientation. We compare to Taxim \cite{si2022taxim} as a baseline, which is implemented on CPU and is the state-of-the-art approach for simulating tactile image.
    \item \textbf{Tactile Force Field}. We use the tactile ball-rolling experimental setup introduced in \cite{xu2023efficient} to evaluate the speed of our simulator in generating tactile normal and shear force fields at varying resolution levels. We compare our results to \cite{xu2023efficient} as a baseline.
\end{enumerate}

Table~\ref{table:tactile_rendering_speed} shows that the effective speed of our simulator increases as we increase the number of simulated parallel environments before the GPU gets saturated. Our simulator runs at $1631$ frames per second (FPS) with $512$ parallel environments on an NVIDIA RTX 3090, compared to the baseline (Taxim \cite{si2022taxim}) that runs at $7.28$ FPS on a single core of an AMD Threadripper 1950X processor. This represents a more than $200\times$ speed-up by \textbf{TacSL} over a baseline with comparable tactile-image quality. 
This significant performance improvement is attributed to \textbf{TacSL} full utilization of the GPU-based parallel processing capabilities available in Isaac Simulator, with most computations performed on the GPU and minimal data transfer overhead between the GPU and CPU.

For tactile shear force fields, we compare the simulation speed at two tactile force field resolutions, $10\times 10$ and $100\times 100$. The $10\times 10$ resolution captures the typical tactile shear resolution used by most real-world visuotactile sensors (e.g. GelSight), providing a speed evaluation for practical scenarios. On the other side, $100\times 100$ represents a stress test of the simulation's speed on a high-resolution tactile sensing field. Table~\ref{table:tactile_field_speed} shows that for low-resolution tactile field ($10\times 10$), the speed of our simulator linearly scales up with the number of environments. When we use $32768$ parallel environments for the ball-rolling task, our simulator achieves $1541043$ FPS which is $428\times$ speedup over the CPU baseline \cite{xu2023efficient}. For the high-resolution stress test case ($100\times 100$), the speed of our simulation gets saturated when we increase the number of parallel environments to $4096$. With $4096$ parallel environments, our simulator is able to simulate $100\times 100$ tactile force field in $103493$ FPS, which is still $46\times$ speedup over the baseline. 

To further analyze the simulation speed, we break down each step of our simulation into two phases: physics simulation and tactile computation. 
The physics simulation phase evolves the state of the system based on laws of physics, while the tactile computation phase calculates the tactile sensor signals (i.e., tactile images and tactile force field) from the current state of the system.
In this analysis, we use $512$ parallel environments for simulating the tactile image setup, and a $10\times 10$ resolution  with $32768$ parallel environments for the tactile force field setup. Table \ref{table:speed_breakdown} reports the total time spent (per environment) and the breakdowns for each simulation step in both tactile modality simulations. As shown in the table, the majority of simulation time is spent on the tactile computation phase after significant speed-ups by TacSL.
By leveraging GPU parallelization, TacSL reduces the simulation time for each tactile modality, consequently decreasing the overall computation time.

\begin{table*}%
\centering
\caption{\textbf{Tactile Rendering Speed.} We compare the rendering speed of TacSL (\textit{OURS}) against a recent baseline \cite{si2022taxim}. Our solution is not only faster for a single environment, but more importantly, is highly parallelized, allowing substantial speed-ups at 512 parallel environments (224x relative to \cite{si2022taxim}). Unit: FPS (frames per second).} 
\label{table:tactile_rendering_speed}
\begin{tabular}{c|c|cccccccccc}
\hline
         & \cite{si2022taxim} & \multicolumn{10}{c}{OURS}                                                                         \\ \hline
Num Envs & 1         & 1      & 2      & 4       & 8       & 16      & 32      & 64     & 128      & 256       & 512     \\ \hline
FPS      & 7.28      & 140    & 259    & 456     & 740     & 1045    & 1288    & 1470   & 1545     & 1609      & 1631 \\ \hline
\end{tabular}
\end{table*}

\begin{table*}%
\centering
\caption{\textbf{Tactile Force-Field Generation Speed.} We compare the force-field generation speed of our simulator (OURS) at two levels of parallelization against a recent baseline \cite{xu2023efficient}. N/A corresponds to scenarios where the data does not fit on the single GPU setup used for our evaluation. Unit: FPS (frames per second).}
\label{table:tactile_field_speed}
\begin{tabular}{cc|c|ccccccccc}
          & & Baseline & \multicolumn{9}{c}{OURS}                                               \\ \hline
& Num Envs  & 1        & 1   & 4   & 16   & 64    & 256   & 1024   & 4096   & 16384   & 32768   \\  \hline
\multirow{2}{*}{Resolution} & 10 x 10   & 3596     & 152 & 734 & 2923 & 10162 & 32892 & 127932 & 452372 & 1287686 & 1541043 \\
& 100 x 100 & 2246     & 144 & 530 & 2131 & 8070  & 27106 & 65213  & 103493 & N/A     & N/A     \\ \hline
\end{tabular}
\end{table*}

\begin{table*}%
\centering
\caption{\textbf{Breakdown of Simulation Time.} Our simulation has two phases: the physics simulation phase and the tactile computation phase. This table reports the breakdown time for a single-step simulation. The time is also divided by the number of parallel environments to represent a per simulation environment step speed. For tactile image generation, we report the time for $512$ parallel environments. For force field generation, we report the simulation time for $10\times 10$ tactile field with $32768$ parallel environments.}
\label{table:speed_breakdown}
\begin{tabular}{c|ccc}
Tactile Modality & Physics Simulation & Tactile Computation &  Total   \\ \hline
Image & 0.146 $ms$   & 0.467 $ms$  &  0.613 $ms$   \\ \hline
Force Field ($10\times 10$) & 0.188 $\mu s$ &  0.461 $\mu s$  & 0.649 $\mu s$     \\ \hline
\end{tabular}
\end{table*}

\subsection{Policy Learning Results}
\subsubsection{Tasks}
We use the following tasks to evaluate our tactile simulator and policy-learning toolkit:
\begin{itemize}
    \item \textbf{Peg Placement}. Here the robot has to place a cylindrical rod upright onto a flat support surface. Without knowing the precise position or orientation of the rod within the gripper, the robot needs to use its sense of touch to implicitly estimate the pose of the rod and align it to be perpendicular with the support surface. %
    Specifically, the rod in the gripper can be rotated (within $\pm$ 20 degrees) away from being coaxially aligned with the gripper. This makes the task harder by requiring rotational alignment that can only be realizable with the help of tactile sensors.
    \item \textbf{Peg Insertion}. Here the robot has to insert a cylindrical rod into a cylindrical socket\cite{narang2022factory,tang2024automate}. Similar to the Peg Placement task, the rod starts at a random position and orientation within the robot's gripper. 
    \item \textbf{Bolt-on-Nut Alignment for Screwing}.  Here the robot must place a bolt onto the threaded hole of a hex nut such that a screwing motion primitive at the end of the episode results in successful fastening. This task requires the bolt to be aligned with the gripper to have a chance of succeeding. We leave this as a challenge task within the \textbf{TacSL} task suite and report results for the other two tasks.
\end{itemize}%

\begin{figure*}[h]
\includegraphics[width=1.0\textwidth]{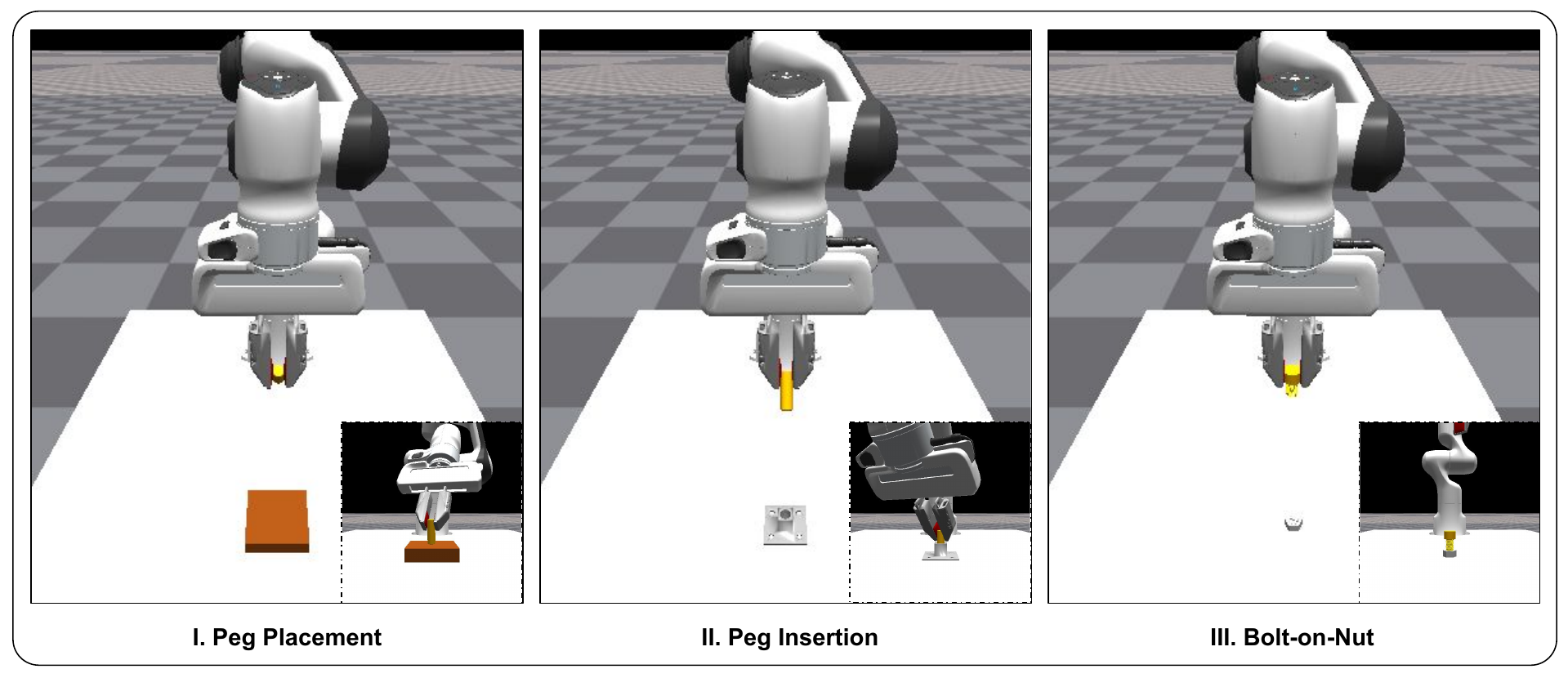}
\caption{\textbf{TacSL Tasks.} Large images show randomized initial configurations of each task, and inset images show final configurations of successful episodes. I) Peg Placement task, where the peg held by the gripper must be placed in an upright pose on a flat plate, such that the peg remains stable upon release. II) Peg Insertion task, where the peg held by the gripper must be inserted into a socket fixed to the table. III) Bolt-on-Nut task, where the bolt held by the gripper must be screwed into a nut fixed to the table, such that the bolt remains stable upon release.}
\label{fig:tasks}
\end{figure*}

\subsubsection{Simulation Results}

We present policy results on one of the tasks discussed above (Peg Insertion) and compare the learning performance across difference sensing modalities. In each case, we show the learning performance for behavior cloning, DAgger, or asymmetric actor-critic RL (AAC), for a combination of different policy inputs.

The policy input options are as follows:
\begin{enumerate}%
    \item Privileged State: Robot arm and gripper joints (9), end-effector pose (7) and velocity (6), socket position (3) and orientation (4), plug position (3) and orientation (4), plug-socket contact force (3), plug-finger-1 contact force (3), plug-finger-2 contact force (3)
    \item Reduced State: Robot arm and gripper joints (9), end-effector pose (7), socket position (3) and orientation (4). Here, a noisy estimate of the socket position is provided, as is common in real world pose estimation. 
    \item High-Dimensional Information: This can be one or more of the following:
    \begin{enumerate}
        \item Tactile image on both fingers: ($2$ x ($80 \times 60 \times 3$))
        \item Tactile Force Field on both fingers: ($2$ x ($14 \times 10 \times 3$))
        \item Wrist camera: ($64 \times 64 \times 3$)
    \end{enumerate}
\end{enumerate}
The action space is a 6D end-effector pose target relative to the current pose. This is used to compute pose targets for a lower-level task-space impedance controller.

\begin{table*}%
\centering
\caption{\textbf{TacSL Policy Learning Results (Simulation).} We compare the policy learning success rates for the peg placement and peg insertion tasks using different sensing modalities. The comparison is performed for three different learning algorithms: offline policy distillation (BC), online policy distillation (DAgger), and asymmetric actor-critic RL (AAC). The observation input to the policies can be either the full privileged state, reduced low-dimensional state available in the real world, or reduced state combined with high-dimensional observations such as tactile images (Tactile-Img), tactile force fields (Tactile-FF), wrist camera images, or a combination. All algorithms perform well on the peg placement task (other than for reduced state), whereas the online algorithms (DAgger and AAC) perform better than offline BC on the peg insertion task.}

\begin{tabular}{ccccccccc}
\toprule
& \textbf{Algorithms} & 
\begin{tabular}[c]{@{}c@{}}Privileged  \\State\end{tabular} &
\begin{tabular}[c]{@{}c@{}}Reduced \\ State\end{tabular} & 
\begin{tabular}[c]{@{}c@{}}Reduced \\ + Tactile-Img \end{tabular} & 
\begin{tabular}[c]{@{}c@{}}Reduced \\ + Tactile-FF \end{tabular} & 
\begin{tabular}[c]{@{}c@{}}Reduced \\ + Wrist \end{tabular} & 
\begin{tabular}[c]{@{}c@{}}Reduced \\ + Tactile-Img-FF \end{tabular} & 
\begin{tabular}[c]{@{}c@{}}Reduced \\ + Tactile-Img \\+ Wrist \end{tabular}\\ 
\midrule

\multirow{3}{*}{\textbf{Placement}} & BC & $0.999$  & $0.319$  & $0.994$  & $0.993$ & $0.995$ & $0.984$  & $0.984$ \\
& Dagger & $1.000$  &  $0.409$ & $0.990$  & $1.000$ & $0.998$ & $0.997$  & $0.990$ \\
& AAC  & $0.999 $  & $0.421$ & $0.999$  & $0.999$  & $0.995$ & $0.999$  & $0.996$ \\
\midrule

\multirow{3}{*}{\textbf{Insertion}} & BC & $0.826$ & $0.087$ & $0.820$ & $0.804$ & $0.936$ & $0.570$ & $\textbf{0.941}$\\
& Dagger & $0.958$  &  $0.053$ & $\textbf{0.908}$  & $0.925$ & $\textbf{0.968}$ & $0.876$  & $0.925$ \\
& AAC  & $\textbf{0.973}$ & $0.0$ & $0.834$ & $\textbf{0.930}$ & $0.948$ & $\textbf{0.932}$ & $0.897$\\
\bottomrule
\end{tabular}
\label{table:sim_policy_results}
\end{table*}

\begin{figure}%
\begin{center}
\includegraphics[width=0.9\linewidth]{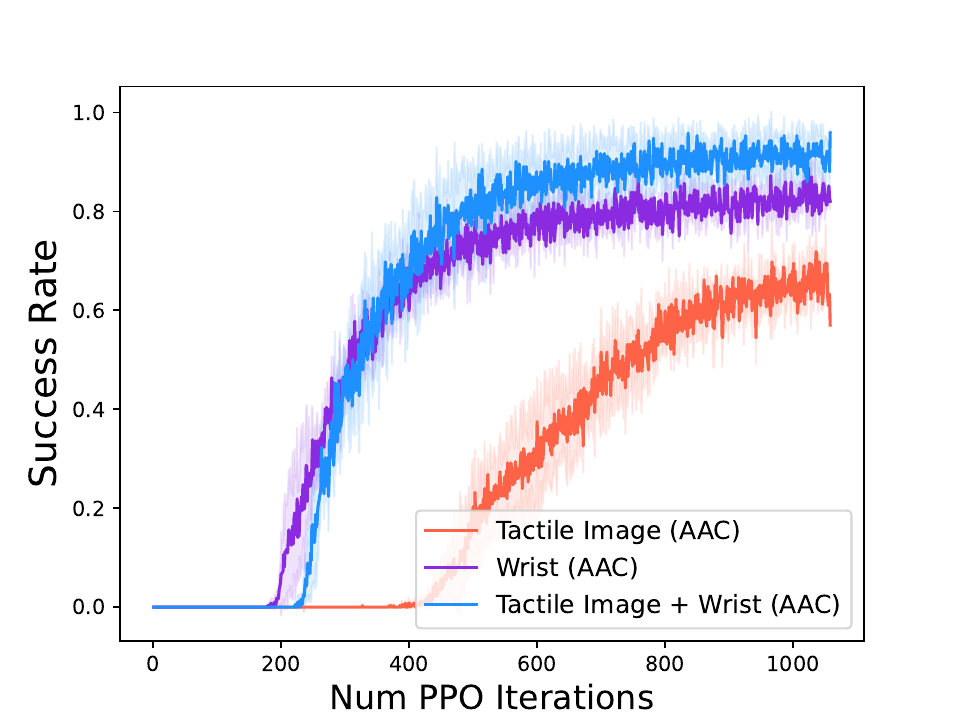}
\end{center}
\caption{\textbf{Multimodal Learning}: \textit{Left}: Comparison of training curves when learning from tactile images, a wrist image, or both. The image input has minimal image augmentation (i.e., channel swapping), and the policy takes in a noisy estimate of socket position ($5$~mm uniform noise).
}
\label{fig:multi_modal_learning}  
\end{figure}

Shown in Table~\ref{table:sim_policy_results}, we report the average success rates of three policies (unique seeds) for each algorithm-input combination. Each evaluation is performed for 1024 randomized initial configurations (i.e., arm configuration, peg-in-gripper position/orientation and insertion/placement location). %
When using privileged state information, the agent is able to learn the task with a high success rate; however, there is a drop in performance without certain states that are difficult to obtain in the real world, such as pairwise contact forces (unavailable) and object-in-gripper pose (difficult to estimate). %
However, both the tactile image and wrist camera inputs recovered high performance on the task, and combining both yielded even higher success rates and learning speeds (Fig.~\ref{fig:multi_modal_learning}).

We hypothesize that the wrist camera and tactile sensors are complementary; while the wrist sensor provides a slightly broader view of the object tip and target location, the tactile sensor focuses on fine-scale in-gripper estimates and serves as a proxy for contact forces. Also, as shown in  Fig.~\ref{fig:multi_modal_learning}, the wrist camera tends to learn faster, as it is able to quickly resolve the socket location given that only a noisy estimate is provided as policy input. On the other hand, the tactile policy relies on the sense of touch, enabling unique robustness to lighting and illumination changes.
For the remainder of our analysis, we focus on transferring tactile image policies from simulation to reality; a comparative analysis of the transferability of other sensor streams is an interesting direction for future work.

\subsubsection{\sysName{} Policy Learning Results}
\begin{figure}[h]
\begin{center}
\includegraphics[width=0.4925\linewidth]{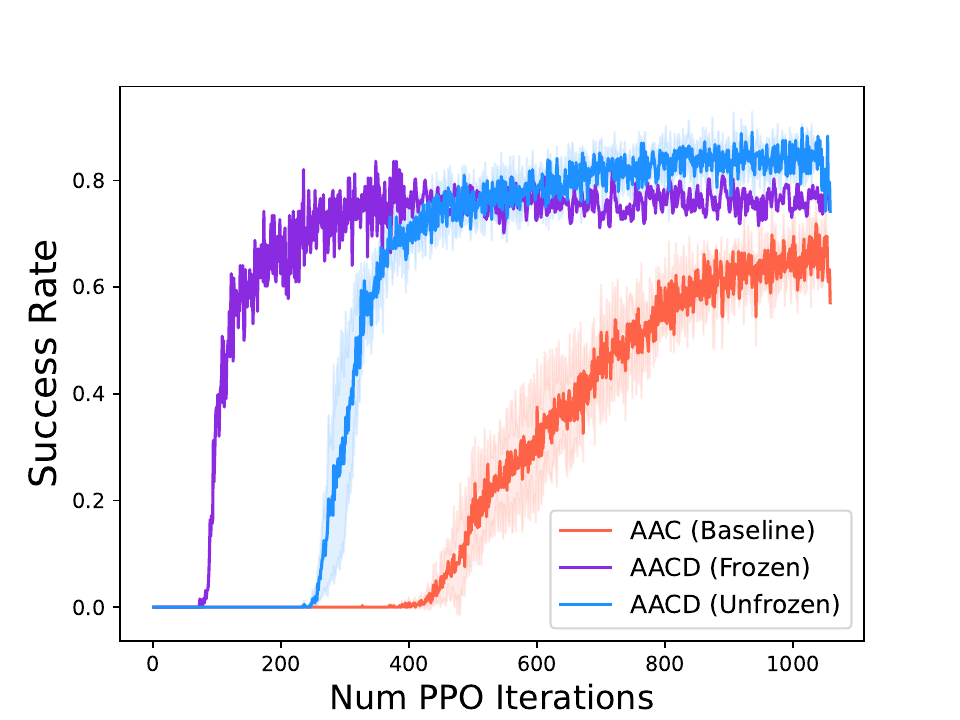}
\includegraphics[width=0.4925\linewidth]{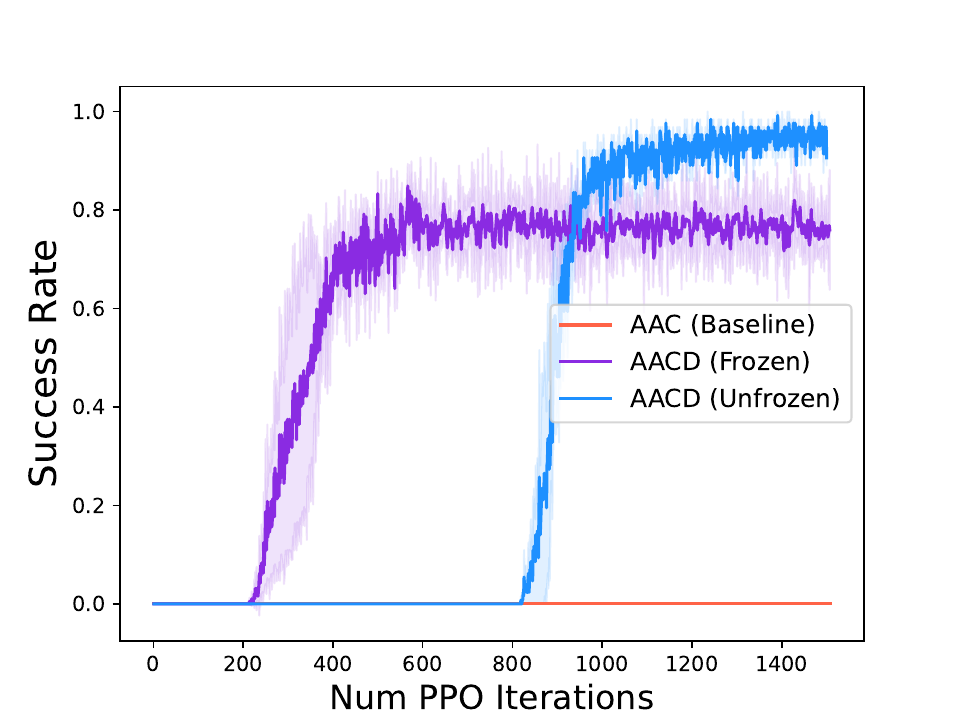}
\end{center}
\caption{\textbf{Effect of Pretrained Critic}:  Training curves for policy learning from tactile images with varying levels of image augmentation. Orange represents a randomly-initialized critic, purple denotes a frozen pretrained critic, and blue signifies an unfrozen pretrained critic. \textit{Left}: minimal image augmentation (channel swapping only). \textit{Right}: full image augmentation needed for sim-to-real transfer.
The plots show that training with a pretrained critic offers two advantages: accelerated training and higher asymptotic performance.
}
\label{fig:aacd_results}  
\end{figure}

In this section, we experimentally evaluate the value of the pretrained critic in our AACD algorithm as compared to the conventional AAC algorithm. Seen in Fig.~\ref{fig:aacd_results}, our results show that a pretrained \textit{expert} critic accelerates policy learning in high-dimensional settings, especially when tactile images serve as input. %
Specifically, training is fastest when the critic is frozen, followed by when the critic is unfrozen; both scenarios are faster than when the critic is initialized with random weights. In addition, the unfrozen critic achieves the highest asymptotic performance, likely because it is capable of acquiring strategies tailored for an image-based agent. %
Furthermore, the benefit of the pretrained critic used in {\sysName} becomes more pronounced with a higher level of image augmentation (see right plot in Fig.~\ref{fig:aacd_results}).
While both versions of {\sysName} successfully learn the insertion task, the baseline (AAC) is unable to learn the task. This result highlights the challenge of performing high-dimensional RL for high-precision contact-rich tasks and demonstrates the effectiveness of utilizing a pretrained critic, as proposed in {\sysName}.

\subsection{Real-Robot Results}
In addition to developing, training and testing our algorithms in simulation, we also tested the trained policies on the real robot via zero-shot sim-to-real transfer. 

\begin{figure}%
\includegraphics[width=1.0\linewidth]{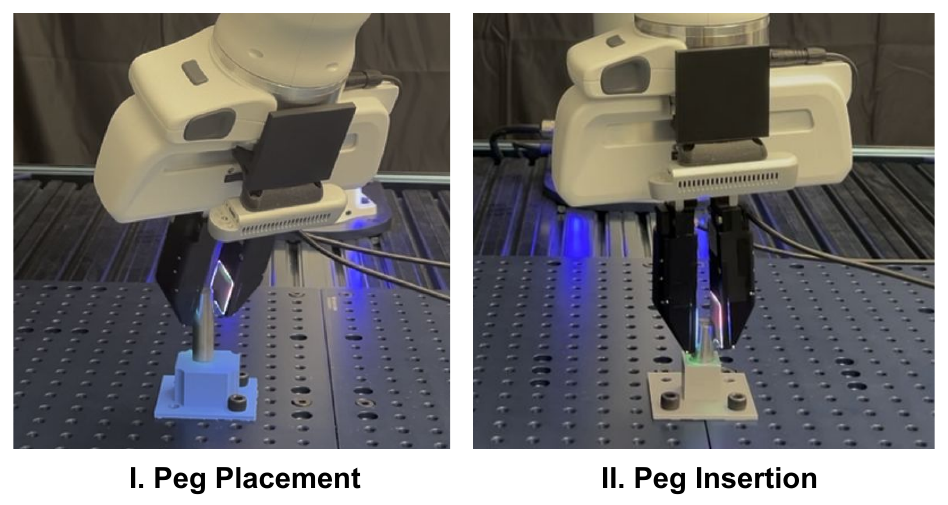}
\caption{\textbf{Policy Deployment in the Real World:} Left: Peg Placement task, where the robot orients and places a $16$~mm-diameter peg upright on a flat blue pedestal. Right: Peg Insertion task, where the robot orients and inserts the peg into a grey socket with a diametral clearance of $5$~mm. The socket is a higher-clearance version of an asset from \cite{tang2023industreal}}
\label{fig:real_robot_setup}
\end{figure}

\paragraph{Peg Placement Task}
We extensively evaluated the peg placement policies at different placement locations. (See experiment setup in Figure~\ref{fig:real_robot_setup}.) For each placement location, we conducted 27 different runs, varying the peg-in-gripper position, peg-in-gripper orientation, and initial pose of the robot end-effector relative to the peg placement location.

Given that tactile sensors degrade over time, requiring periodic replacement of the elastomeric surfaces due to wear and tear, it is crucial for the trained policy to exhibit robustness to variations such as tactile stiffness and colorations %
To achieve this, the policies were trained with domain randomization strategies described in Section \ref{sec:bridging_sim2real}. Additionally, we explored three different ways of representing tactile input to the policy:
\begin{itemize}
    \item \textbf{Color}: This uses the raw tactile RGB image.
    \item \textbf{Diff}: This takes the difference between current tactile image and the nominal  tactile image. The nominal tactile image is the measurement taken when no object is touching the sensor.
    \item \textbf{Concat}: This concatenates the current tactile and nominal measurement to obtain a 6-channel input.
\end{itemize}
We trained and evaluated two versions of the Color representation. \textit{Vanilla} was trained on a well-calibrated simulated RGB tactile sensor and tested on a different set of real sensors. \textit{ColorAug} was trained with the image augmentation method described in Section \ref{sec:tactile_image_aug}. \textit{Diff+ColorAug} and \textit{Concat+ColorAug} are the Diff and Concat input types, respectively, that were also trained with image augmentation. 

\begin{figure}[h]
\includegraphics[width=1.0\linewidth]{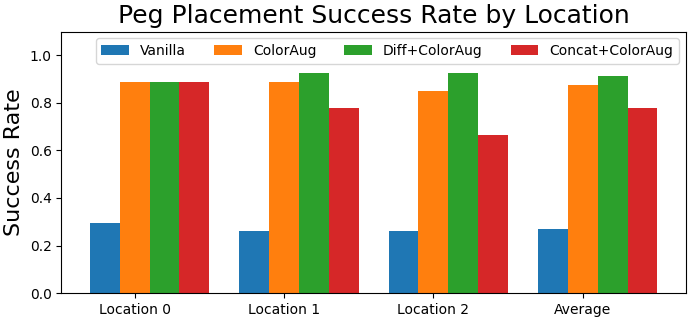}
\caption{\textbf{Peg Placement Results (Real)} We evaluate the zero-shot peg placement task performance (success rate) using policies trained with different representation of the tactile images on the real robot. \textbf{Vanilla}: trained on a single calibrated raw tactile RGB image. \textbf{Color-Aug}: additionally trained with image augmentation. \textbf{Diff+ColorAug}: utilizes the difference between the current tactile image and the nominal tactile image.  \textbf{Concat+ColorAug}: concatenates current and nominal tactile images to form a 6-channel policy input. 
Each policy was evaluated at three different placement locations. For each placement location, we %
randomized the initial end-effector pose, peg-in-gripper position, and peg-in-gripper rotation $\{-\pi/6, 0, \pi/6\}$.
Note that the performance is consistent across placement locations.
The average success rate is shown on the right bar plot.}
\label{fig:real_robot_results}
\end{figure}

Figure~\ref{fig:real_robot_results} shows that the image augmentation and other strategies described in Section \ref{sec:bridging_sim2real} enabled zero-shot sim-to-real policy transfer on the peg placement task. The \textbf{ColorAug} policy with raw RGB inputs trained with image augmentation achieved an average of $87.7\%$ across 81 different trials compared to $27.2\%$ achieved by the \textbf{Vanilla} policy which is not trained with image augmentation.
Empirical results revealed that, by taking the difference between the current tactile image and the nominal tactile image, \textbf{Diff+ColorAug} achieved a $91.4\%$ success rate. On the other hand, by allowing the network to autonomously discover operations to combine the current and nominal tactile images, \textbf{Concat+ColorAug} achieved $77.9\%$.
We hypothesize that our simple network architecture limited the network's ability to discover a better image operation, and we suspect that employing a larger and more sophisticated network architecture could enable the agent to learn improved ways of combining them. We leave this as an interesting avenue for future work.

\paragraph{Peg Insertion Task}
We also demonstrated that the insertion policies transfer to the real-world robot. Here, we evaluated the zero-shot task performance (success rate) using a \textbf{Color-Aug} insertion policy. %
We evaluated the policy at three different socket locations, randomizing the initial end-effector pose, peg-in-gripper position and peg-in-gripper rotation ($\{-\pi/12, 0, \pi/12\}$) for a total number of 81 trials. The policy succeeded 67 times, achieving an $82.7\%$ success rate without any additional real-world fine-tuning. 

Importantly, both the placement and insertion policies exhibited reactivity and robustness to human perturbation of the peg-in-gripper during policy execution (see videos on the project \href{https://iakinola23.github.io/tacsl}{website}).
We show how tactile sensing can be effective in handling reflective metallic parts in challenging lighting conditions that are typical in industrial settings.
To the best of our knowledge, this is the first work to show a reactive, robust tactile policy that can withstand significant in-gripper object perturbation during execution of precise manipulation tasks.

\section{Limitations and Future Work}
Our work has several limitations, which present opportunities for future work.
First, our soft contact model uses a Kelvin-Voigt constitutive law, which is linear. However, there are nonlinear variants that may provide improved accuracy, such as one proposed by Hunt and Crossley \cite{hunt1975coefficient, marhefka1999compliant}. 
Second, in order to map RGB to depth, we use a calibrated look-up table. However, a more accurate mapping may be achieved through a learned model~\cite{chen2022bidirectional}, especially when handling curved sensors\cite{gomes2020geltip,tippur2023gelsight360}. Third, while \textbf{TacSL} focuses on transferring tactile image policies from simulation to real, it would be interesting to also integrate policy transfer for other modalities such as the force-field modality, as demonstrated in prior work\cite{xu2023efficient}.
Finally, \textbf{TacSL} combines accelerated simulation with standard recurrent neural network architecture to achieve strong baselines for tactile policy learning. Leveraging other architectures such as transformers and diffusion models is an avenue for future work.

\section{Conclusion}
\label{sec:conclusion}

We have presented \textbf{TacSL}, an accelerated tactile simulator that gives both geometric and force field information. \textbf{TacSL} includes a suite of contact-rich tasks and a toolkit of online learning algorithms for tactile policy learning, including a novel RL algorithm (\sysName) that enables efficient policy learning in high-dimensional domains. Using \textbf{TacSL}, we analysed the performance of tactile and other sensing modalities in solving the peg-placement and peg-insertion task in simulation. Furthermore, \textbf{TacSL} prescribes tactile policy-training strategies that transfer in zero-shot to the real world. Once released, we believe our simulation and policy learning framework will be a highly-useful testbed for leveraging tactile sensing for a wide range of contact-rich robotic tasks.

\section*{Acknowledgments}
We would like to thank our colleagues at NVIDIA for their invaluable assistance and feedback throughout this work. Special acknowledgment to Michael Noseworthy, Bingjie Tang, Bowen Wen, Karl Van Wyk, Ankur Handa and Fabio Ramos for engaging in deep conversation and providing insightful feedback. We deeply thank Philipp Reist for his insights on the physics solver and detailed feedback on the paper draft. We appreciate the assistance of Tobias Widmer and Milad Rakhsha in the SDF implementation. Our thanks also go to Viktor Makoviychuk and Kelly Guo for their responsiveness to our inquiries about Isaac Simulator. 
We thank Ajay Mandlekar for assistance in setting up behavior cloning during early prototyping, and Alperen Degirmenci for help in creating visualizations and figures for the paper.
Valuable feedback on the paper draft was provided by Yu-Wei Chao. 
We express our thanks to Kimo Johnson for important GelSight hardware support.

{\appendices
\section{Dynamics Solver Details}
\label{sec:appendix_dyn_solver}

\textbf{TacSL} employs the Temporal Gauss-Seidel (TGS) algorithm, implemented in the NVIDIA PhysX SDK~\cite{PhysX}, as its dynamics solver. The TGS solver is an advanced iterative method designed for efficiently and robustly resolving constraints in systems with complex interactions, such as collisions, friction, and joints.
The solver divides each frame duration into $N$ substeps, reducing the integration timestep to $\Delta t/N$. During each substep, constraints are processed sequentially.

\paragraph{Constraint Resolution}

For each constraint, such as a collision or joint limit, the TGS solver calculates an impulse $\lambda$ to minimize the constraint error. This impulse is computed using the constraint's gradient and its compliance (if any), incorporating temporal stabilization terms specific to TGS.
These stabilization terms address errors from discrete timesteps by incorporating positional corrections  accumulated over previous time steps into the velocity update~\cite{macklin2019small}, thereby preventing error accumulation over time. 
The computed impulse is then applied to the bodies affected by the constraint by adjusting their velocities $\Delta v$. The sequential nature of the Gauss-Seidel approach ensures that the most recent updates to positions and velocities are used for subsequent constraints, which promotes convergence and enhances stability in tightly coupled systems.

\paragraph{Integration}

After processing all constraints for a substep, the solver integrates the bodies' positions using the updated velocities. This substep integration maintains consistency between positions and velocities throughout the simulation. The combination of substepping and TGS's stabilization minimizes errors from discrete timesteps and improves the handling of dynamic interactions.

For further implementation details, including the source code, we refer the reader to the open-source PhysX SDK repository~\cite{PhysX}.

\section{Contact Model Calibration}
To calibrate the compliant contact parameter, we placed known standardized calibration weights on the tactile sensor in both real and simulated environments. The compliant stiffness parameter ($\kappa$) is chosen such that the surface area of the tactile impression in the simulation and real-world environments roughly match for different known weights.

\begin{figure}[h]
\begin{center}
\includegraphics[width=1.\linewidth]{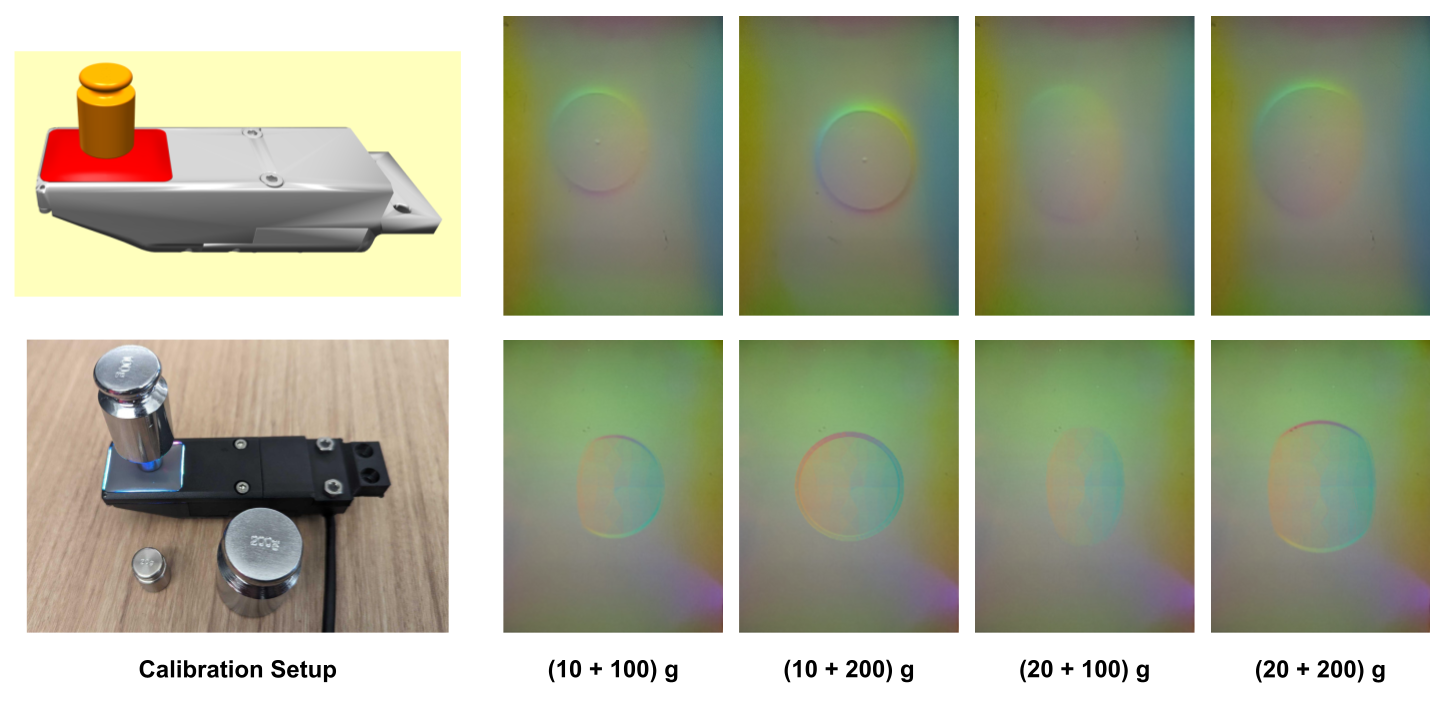}
\end{center}
\caption{\textbf{Contact-model calibration}. Top: Simulated. Bottom: Real. The left column shows images of the calibration setup in sim and real. In real, we stack calibration weights on the sensor, while in sim we place the corresponding geometry and weight on the sensor. The corresponding measurements of the sensor is showed for different weights.
}
\label{fig:contact_model_calib}  
\end{figure}

We use standard weights of 10g, 20g, 100g, and 200g for our calibration. The 10g and 20g weights fit on the GelSight sensor but are too light to make visible tactile impressions. On the other hand, the 100g and 200g weights are heavy enough but too wide to fit on the sensor. Therefore, we combined the two groups by placing the heavy, wide weights on top of the narrow, light weights (see Figure \ref{fig:contact_model_calib}). Since all the weights are known, we obtained four calibration conditions: 110g, 210g, 120g, and 220g.

Figure \ref{fig:contact_model_calib} shows the calibration setup as well as the corresponding tactile readings in the simulation and real-world environments. We found that $\kappa = 200$ worked well for 110g and 210g with a narrow contact surface area of the bottom 10g weight, while 
$\kappa = 300$ worked for 120g and 220g with a wider contact surface area. For policy learning, we selected a domain randomization range of $\kappa = [150-350]$ that encapsulates the values obtained during calibration.

\section{Training Details}
\label{sec:training_details}

Table~\ref{tab:env_rand} shows the task randomization levels when generating initial states for the placement and insertion tasks. The end-effector position is randomly sampled around a home upright pose according to the defined randomization levels, the peg is randomly initialized at a random position and orientation within the gripper, and the socket is randomly placed in front of the robot. Additionally, the soft-contact parameters and robot joint damping values are randomized. Finally, observation noise is added to the location of the socket to reflect imperfect pose estimation of objects in the real world.

\begin{table}[H]
    \centering
    \caption{Environment Randomization Bounds. Each parameter is uniformly sampled within the specified range. Note: the peg-in-gripper randomization ranges are relative to the center of elastomer sensor on the gripper tip.}
    \begin{tabular}{c|c}
    \hline
        Parameter & Randomization range \\
    \hline
        End-effector X-axis (m) & $[0.4 - 0.6]$  \\
        End-effector Y-axis (m) & $[-0.1 - 0.1]$ \\
        End-effector Z-axis (m) & $[0.1 - 0.2]$ \\
        End-effector Euler-X (rad) & $[3.04 - 3.24]$ \\
        End-effector Euler-Y (rad) & $[-0.1 - 0.1]$ \\
        End-effector Euler-Z (rad) & $[-1.0 - 1.0]$ \\
        Socket X-axis (m) & $[0.4 - 0.6]$ \\
        Socket Y-axis (m) & $[-0.1 - 0.1]$ \\
        Socket Z-axis (m) & $[0.0 - 0.02]$ \\
        Peg-in-gripper Z-pos (m) & $[-0.0125 - 0.0125]$ \\
        Peg-in-gripper X-rot (rad) & $[-0.628 - 0.628]$ \\
    \hline
        Socket X-Y-Z observation noise (m) & $[-0.005 - 0.005]$ \\
        Compliance-stiffness ($\kappa$) noise (N/m) & $[150 - 350]$ \\
        Compliance-damping ($c$) noise (N/(m/s))  & $[0.0 - 1.0]$ \\
        Joint-damping noise (N/(m/s)) & $[-1.5 - 1.5]$ \\
    \end{tabular}
    \label{tab:env_rand}
\end{table}

\textbf{Reward function}:
Adapted from \cite{narang2022factory,noseworthy2024forge}, the reward function is given as:
\begin{align*}
    R_{\text{task}} = r_{\text{keypoint}} - r_{\text{action}} - r_{\text{contact}}, 
\end{align*}
where

\begin{itemize}
    \item $r_{\text{keypoint}}$ is the distance between keypoints centered on the peg and keypoints of its target pose on the placement pad/socket, passed through an exponential function.
    \item $r_{\text{action}}$ is a penalty on the policy action.
    \item $r_{\text{contact}}$ is a penalty on the contact forces between the peg and the environment, including the socket and table.
\end{itemize}

\textbf{Robot policy action and control}:
The policy outputs a 6D pose target for the end-effector, with a maximum position displacement of $0.01$ m and  a maximum orientation displacement of $0.05$ rad in each dimension. The pose target is sent to a task-space-impedance controller at 60Hz.

\textbf{Policy structure}: The RL policy consists of a CNN visual encoder, an LSTM, and an MLP. The features from the visual encoder are concatenated together with the other low-dimensional observations (proprioception, socket position and orientation, etc.) and fed to the LSTM followed by a MLP module. Note that the CNN module is not included for the expert policy training that uses only low-dimensional inputs. Table~\ref{tab:architecture} shows the architectural details.

\begin{table}[H]
    \centering
    \caption{RL Policy Structure}
    \begin{tabular}{c|c}
    \hline
        CNN channels & [32, 32, 64] \\
        CNN kernel sizes & [8, 4, 3] \\
        CNN stride & [2, 1, 1] \\
        Final CNN pooling layer & Spatial SoftArgMax \\
    \hline
        RNN type & LSTM \\
        RNN layers & 2 \\
        RNN hidden dims & 256 \\
        MLP hidden sizes & 256, 128, 64 \\
        MLP activation & ELU \\
        
    \end{tabular}
    \label{tab:architecture}
\end{table}

\textbf{Hyperparameters}: The PPO hyperparameters adapted from prior work\cite{tang2023industreal} with minimal modifications are shown in Table~\ref{tab:ppo_params}. The BC and DAgger parameters are shown in Table~\ref{tab:bc_params}

\begin{table}[H]
    \centering
    \caption{PPO Hyperparameters}
    \begin{tabular}{c|c}
    \hline
        PPO clip parameter & $0.2$ \\
        GAE $\lambda$ & $0.95$ \\
        Learning rate schedule & adaptive \\
        Initial learning rate & $1e-4$ \\
        Discount factor $\gamma$ & $0.99$ \\
        Critic coefficient & $2.0$ \\
        \# environments & $128$ \\
        \# environment steps per training batch & $512$ \\
        Mini-batch size & $512$ \\
        Learning epochs per training batch & $4$ \\
    \end{tabular}
    \label{tab:ppo_params}
\end{table}

\begin{table}[H]
    \centering
    \caption{BC and Dagger Hyperparameters}
    \begin{tabular}{c|c|c}
                    & BC & Dagger \\
    \hline
        Learning rate & $3e^{-4}$ & $3e^{-4}$ \\
        Number of rollouts per iteration & N/A & $256$ \\
        Number of training epochs per iteration & N/A & $1$ \\
        Sample sequence length & $20$ & $20$ \\
    \end{tabular}
    \label{tab:bc_params}
\end{table}

\subsection{Policy Learning with different assets.}
\textbf{TacSL} can be used to train policies for different assets.
To show this, we trained peg-placement and peg-insertion policies for pegs of three different sizes ($m8$, $m12$, and $m16$) using the same training code and parameters. 
We trained a policy on each individual asset and also trained a fourth policy using all three assets. Shown in Table \ref{tab:policy_asset_sizes}, the results suggest value in training a policy on all assets simultaneously, as policy performance improved across assets, especially for the more challenging ones (the thinnest peg in this case). For both placement and insertion tasks, the policy performance improved especially on the two smaller pegs ($m8$ and $m12$) peg when using the generalist policy trained all three peg sizes compared to the specialist policy trained only on one asset.

\begin{table}[H]
    \centering
    \caption{Policy Learning with Varying Assets: We train specialist policies on each individual asset sizes, and trained a generalist policy on all assets together. The specialist policies were evaluated on the assets they were trained on, and the generalist policy evaluated on each of the three assets. We report the average success rates over 512 trials.}
    \begin{tabular}{cc|cccc}
                    && m8 & m12  & m16\\
    \hline
        \multirow{2}{*}{\textbf{Placement}} & Specialist & $0.824$ & $0.992$ & $\textbf{1.00}$\\
        & Generalist & $\textbf{0.887}$ & $\textbf{0.996}$ & $0.991$\\
    \hline
        \multirow{2}{*}{\textbf{Insertion}} & Specialist & $0.758$ & $0.828$ & $0.883$\\
        & Generalist & $\textbf{0.805}$ & $\textbf{0.906}$ & $\textbf{0.922}$\\
    \hline
    \end{tabular}
    \label{tab:policy_asset_sizes}
\end{table}

\section{Tactile Force Field Edge-Test}

The \textit{edge-test} experiment was designed to validate the shear force-field computation \cite{xu2023efficient}. In this experiment, the robot holds a known cylindrical peg and interacts with the four edges of the corresponding socket. The normalized tactile flow maps collected both in simulation and in the real world are visualized in Figure \ref{fig:edge_test}. Comparing the simulated and real sensor readings, the visualizations show that the directionality of the measurements match and the magnitudes correlate well.

\begin{figure}
  \centering
  \begin{tabular}{@{}c@{}}
    \includegraphics[width=.45\linewidth]{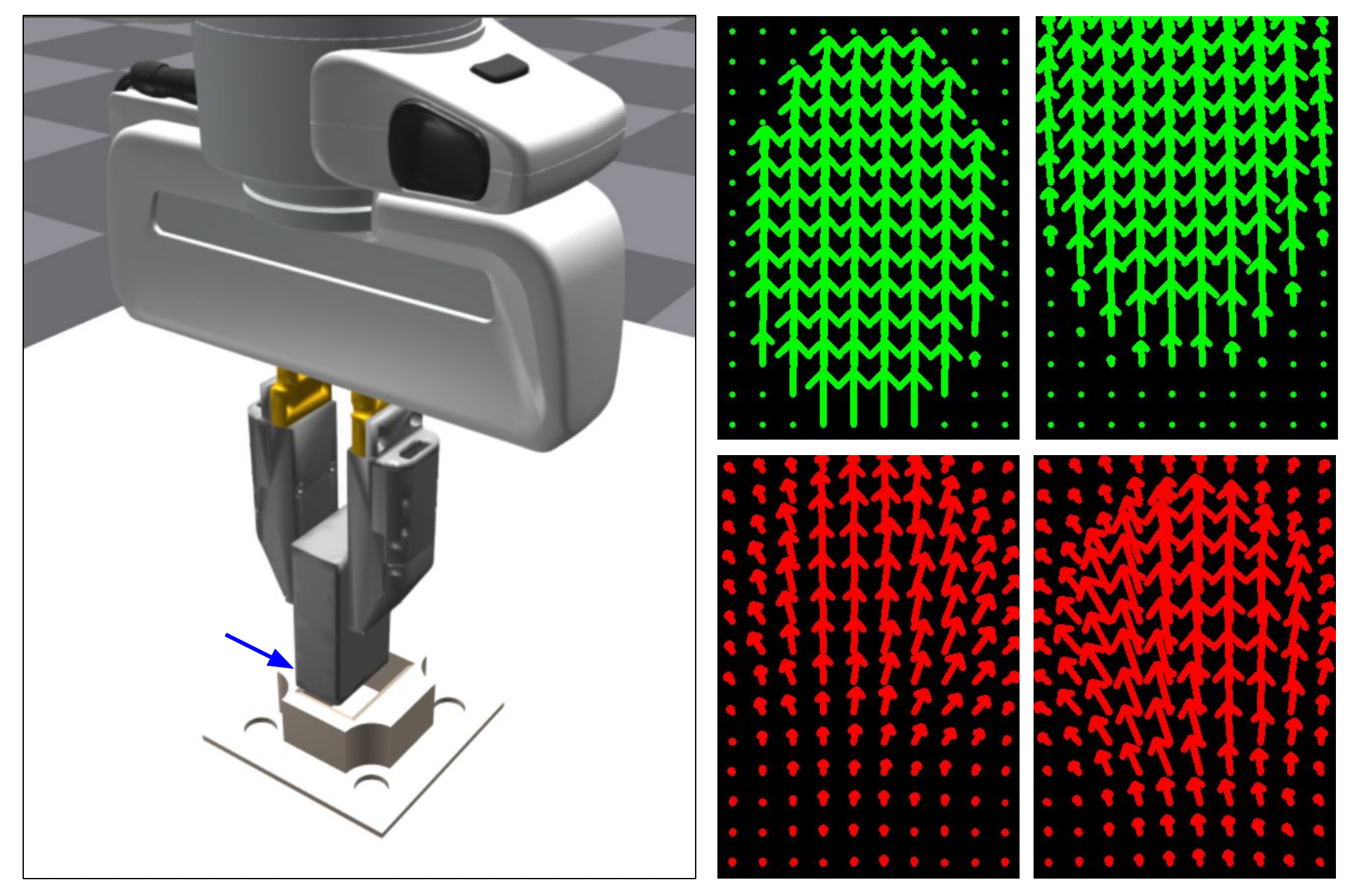} \\[\abovecaptionskip]
    \small (a) Contact at left edge
  \end{tabular}
  \begin{tabular}{@{}c@{}}
    \includegraphics[width=.45\linewidth]{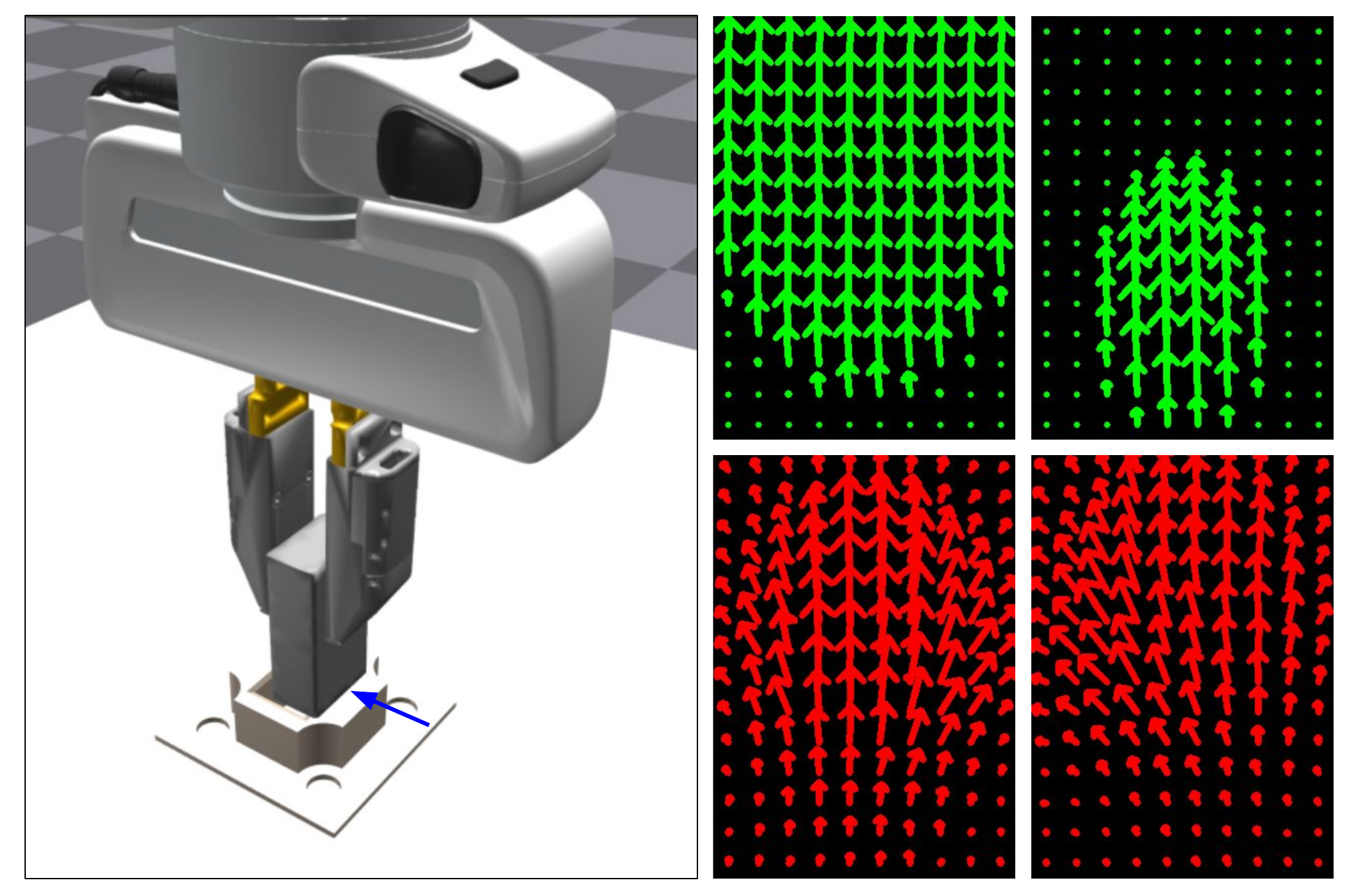} \\[\abovecaptionskip]
    \small (b) Contact at right edge
  \end{tabular}

  \vspace{\floatsep}

  \begin{tabular}{@{}c@{}}
    \includegraphics[width=.45\linewidth]{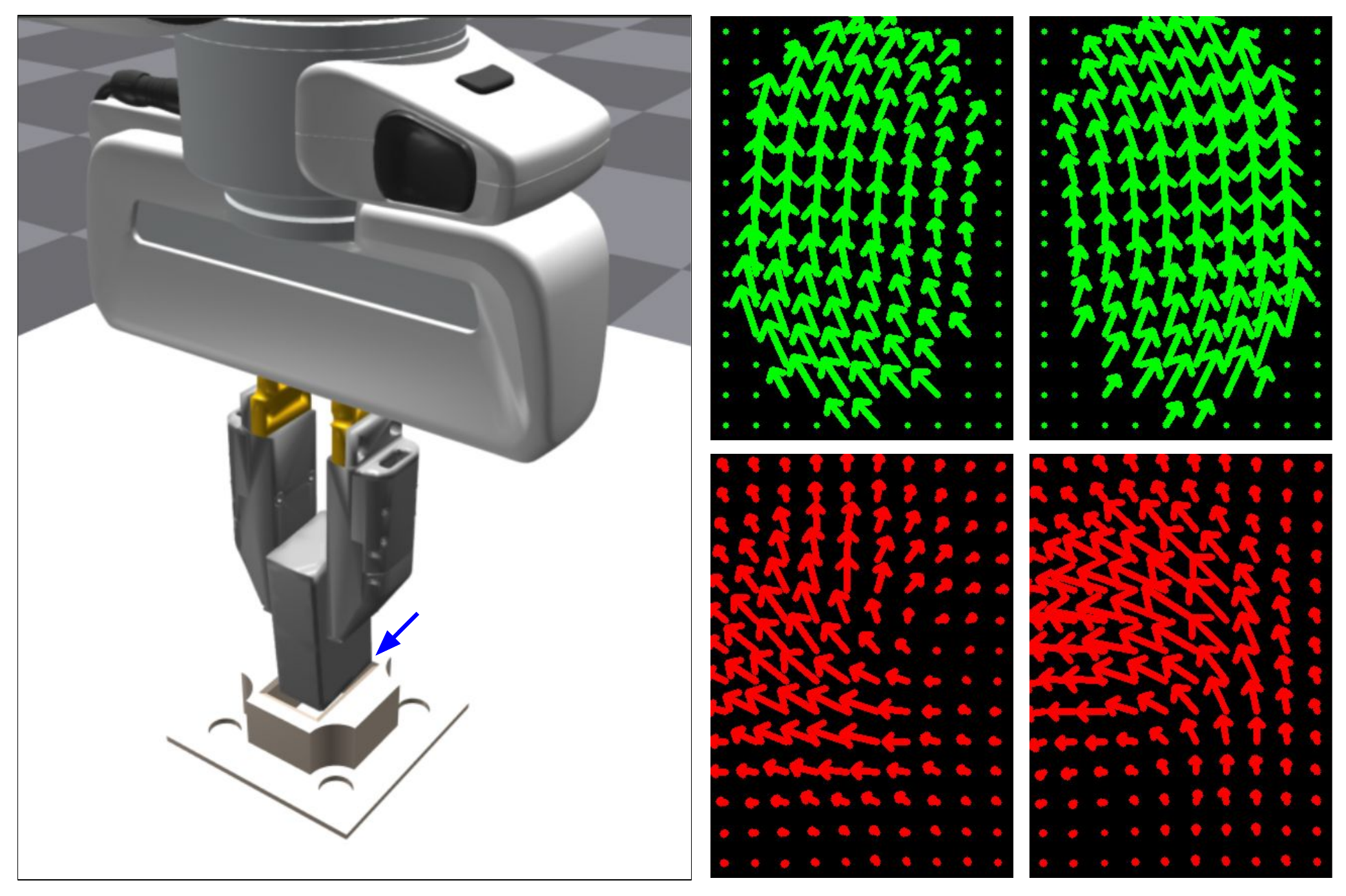} \\[\abovecaptionskip]
    \small (a) Contact at top edge
  \end{tabular}
  \begin{tabular}{@{}c@{}}
    \includegraphics[width=.45\linewidth]{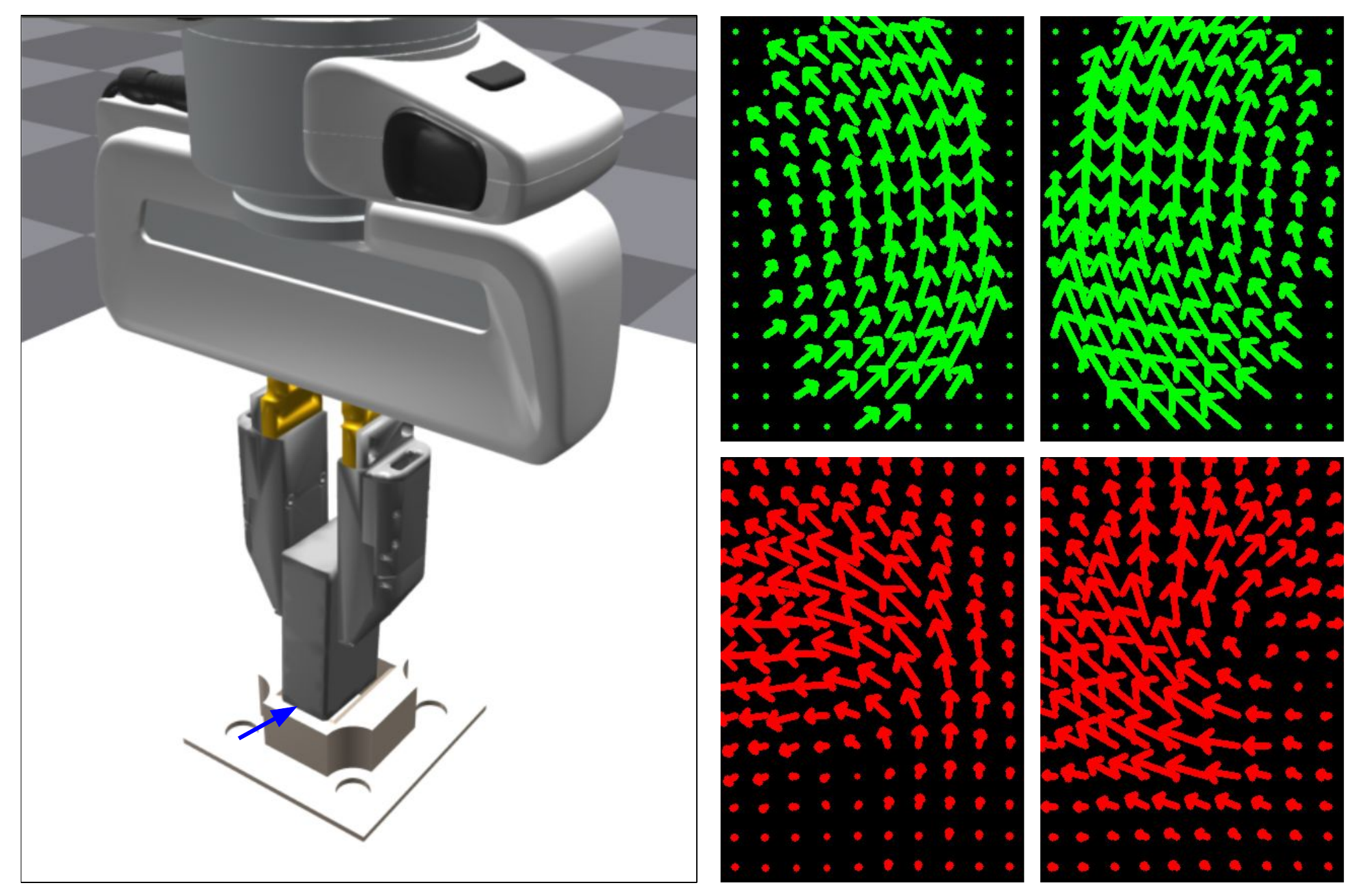} \\[\abovecaptionskip]
    \small (b) Contact at bottom edge
  \end{tabular}

\caption{\textbf{Comparison of normalized tactile force-fields}. In each case, the simulated shear force-fields (green) of the two fingers are at the top, while the real GelSight R1.5 shear force-fields (red) are at the bottom.
}
\label{fig:edge_test}
\end{figure}

\section{Tactile force-field notation}
The tactile force-field computation is a separate post-processing step after each dynamics simulation step. In principle, both the parameter $\epsilon$, used in the formulation of the dynamics solver of the full simulator (Section \ref{sec:contact_simulation}), and the parameter $d$, used in the penalty-based computation of tactile force-field (Section \ref{sec:force_field_simulation}), refer to the same quantity-- penetration depth. However, they refer to the interpenetration depth of different categories of points. The dynamics solver generates contact points based on objects in the scene when resolving collision constraints, while the tactile field computation is done on predefined tactile points which are a sampled grid on the tactile sensor and correspond to the markers that are embedded in the physical sensor.

\section{Tactile Image Visualization}

Fig.~\ref{fig:sim_vs_real_r15} shows a qualitative comparison of a selection of simulated tactile readings and real-world tactile readings. We can achieve maximal sim-real similarity between the real and simulated sensor with careful calibration. However, in practice, sensors degrade over time, and we periodically change the sensor elastomer. To ensure the policy's effectiveness beyond calibrated sensors and its robustness to sensor degradation over time, randomization remains crucial during policy training.

\begin{figure}[h]
\begin{center}
\includegraphics[width=1.\linewidth]{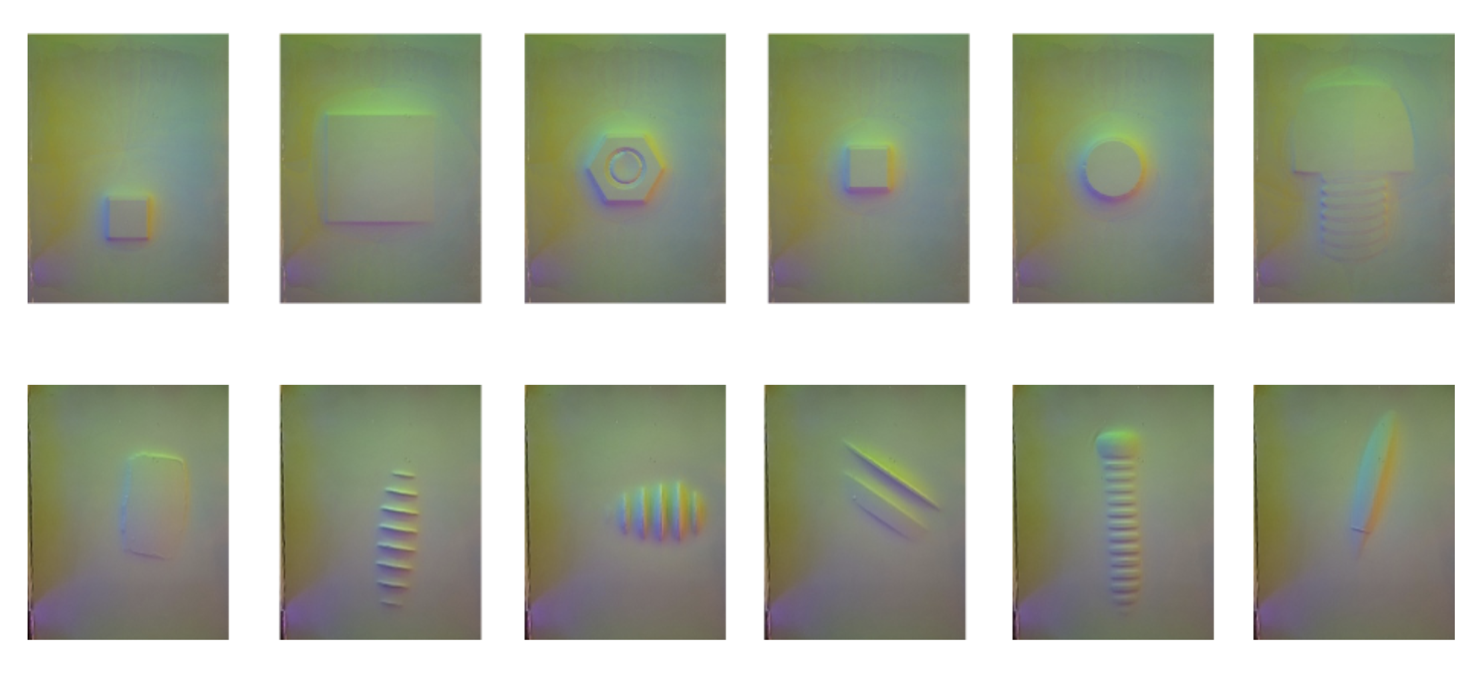}
\end{center}
\caption{\textbf{Simulated and Real-World Tactile Readings}. Top: Simulated. Bottom: Real.
}
\label{fig:sim_vs_real_r15}  
\end{figure}

\textbf{TacSL} can be configured for multiple sensors. Fig.~\ref{fig:multiple_sensors} shows tactile simulation for two GelSight sensors: R1.5 and mini.
Notably, to simulate a given visuotactile sensor, \textbf{TacSL} requires the following:
\begin{itemize}
    \item Elastomer mesh models: A surface mesh model of the thin surface ($0.5$~mm) of tactile sensor serves as the visual mesh, enabling the visualization of the geometry of rigid objects interacting with the sensor. The full volumetric mesh of the elastomer (i.e., not just the thin shell) is used as collision mesh for physics simulation.
    \item Soft contact parameters: These include stiffness and damping parameters that determine the contact interaction between the soft sensor and other rigid bodies. 
    This can be heuristically determined by adjusting the parameters so that the tactile imprint of an object in simulation roughly matches the imprint area observed on the corresponding real tactile image.
    \item Tactile camera pose: This specifies the position and orientation of the camera placed at the back of the sensor.
    \item Tactile camera intrinsic parameters: This includes parameters such as focal length, field of view, and image size.
\end{itemize}
With these specifications, \textbf{TacSL} can simulate tactile depth maps. To convert from depth maps to RGB images, \textbf{TacSL} uses a tensorized calibrated polynomial look-up table obtained through the calibration procedure described in Taxim\cite{si2022taxim}.

\begin{figure}[h]
\begin{center}
\includegraphics[width=1.\linewidth]{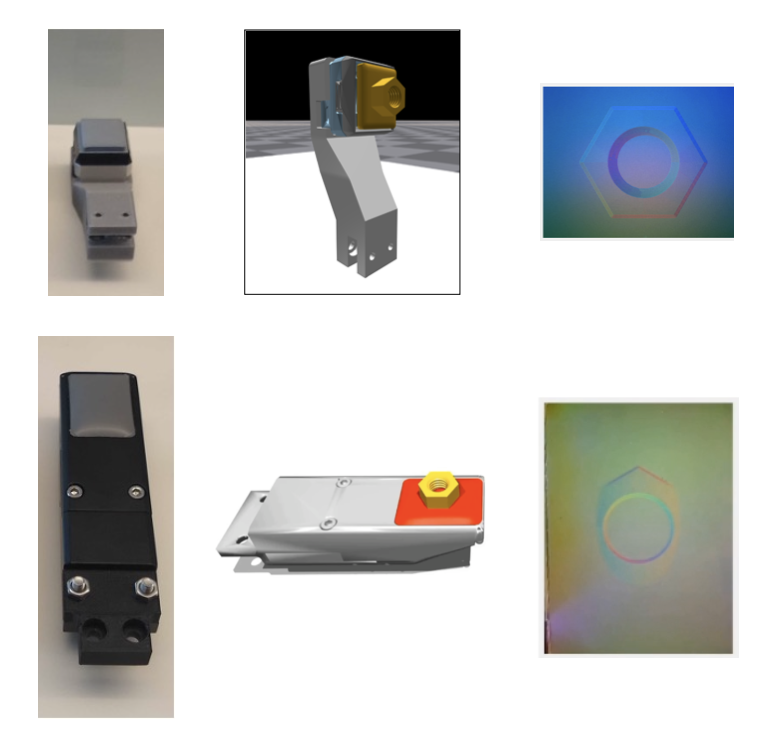}
\end{center}
\caption{
\textbf{Simulation of Multiple Tactile Sensors}. Top row: GelSight Mini. Bottom row: GelSight R1.5.
From left to right: real sensor, simulated sensor, and simulated sensor image.
}
\label{fig:multiple_sensors}  
\end{figure}

\section{Additional Tactile Visualizations}

We present additional visualizations (Figure \ref{fig:additional_viz}) of both tactile images and tactile force fields for the GelSight R1.5 sensor interacting with a complex bolt mesh, which comprises 30k faces and 26k vertices, at various poses.

\begin{figure}%
\begin{center}

  \begin{tabular}{@{}c@{}}
    \includegraphics[width=0.475\linewidth]{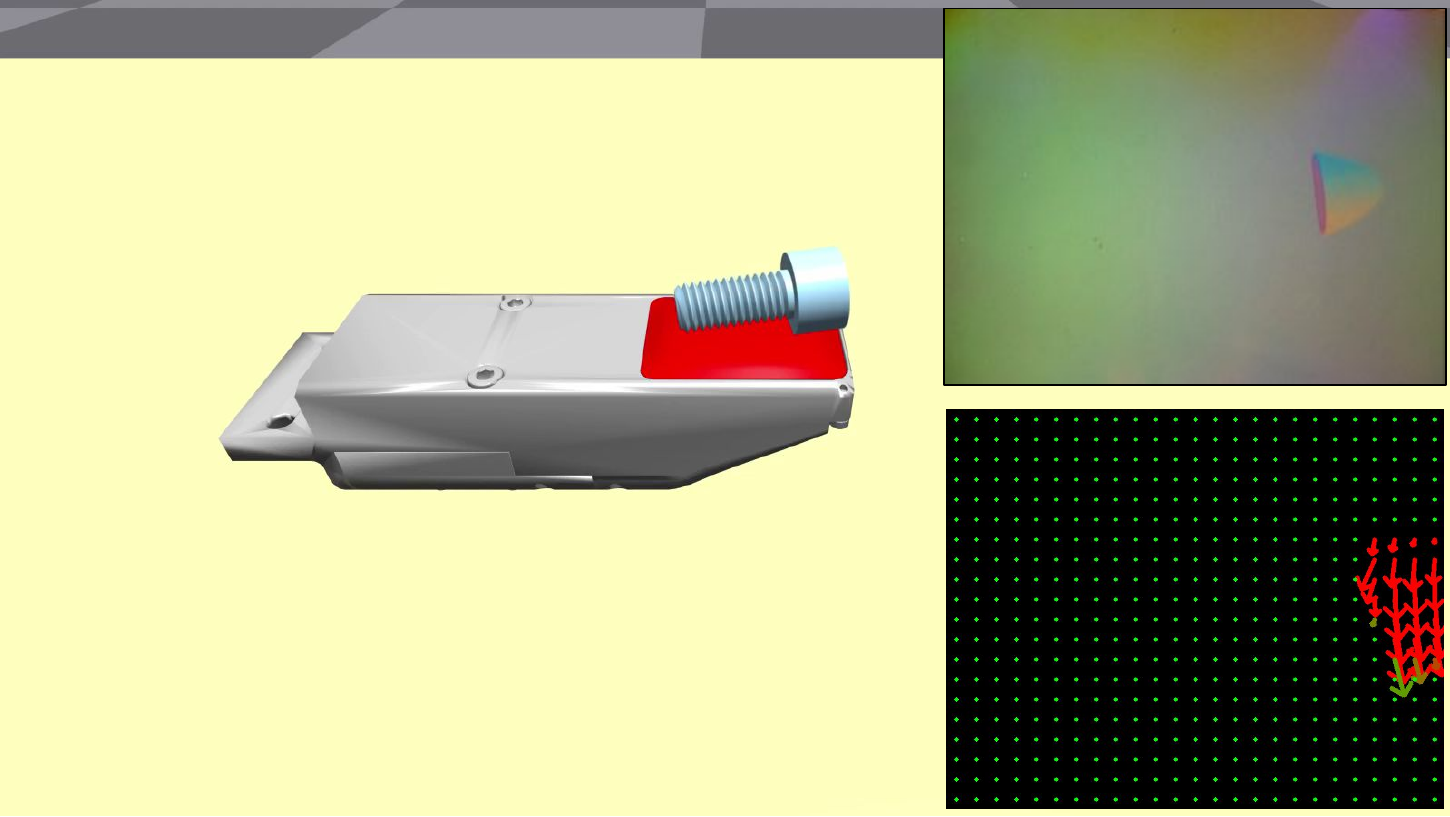} \\[\abovecaptionskip]
  \end{tabular}
  \begin{tabular}{@{}c@{}}
    \includegraphics[width=0.475\linewidth]{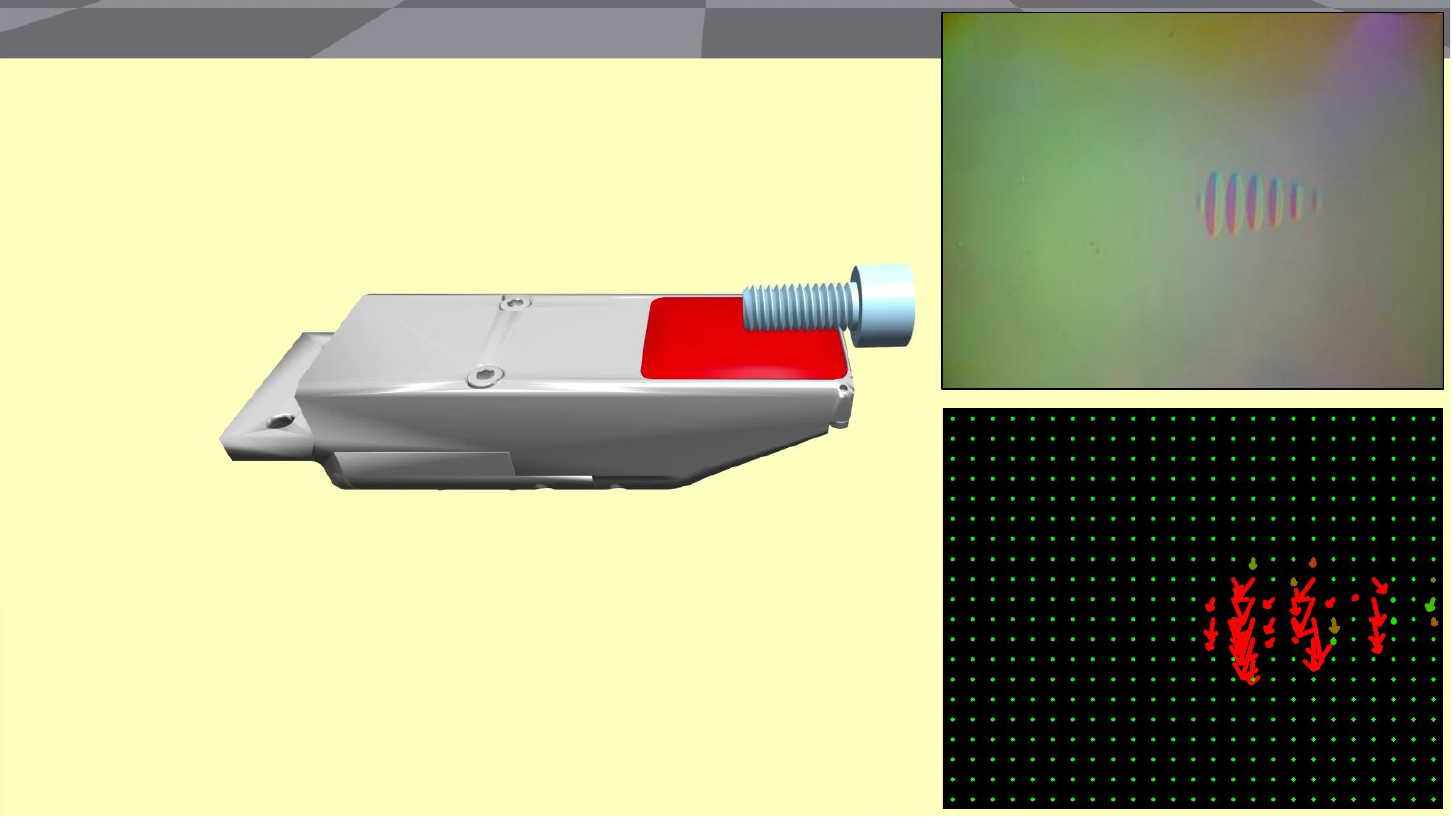} \\[\abovecaptionskip]
  \end{tabular}
  \begin{tabular}{@{}c@{}}
    \includegraphics[width=0.475\linewidth]{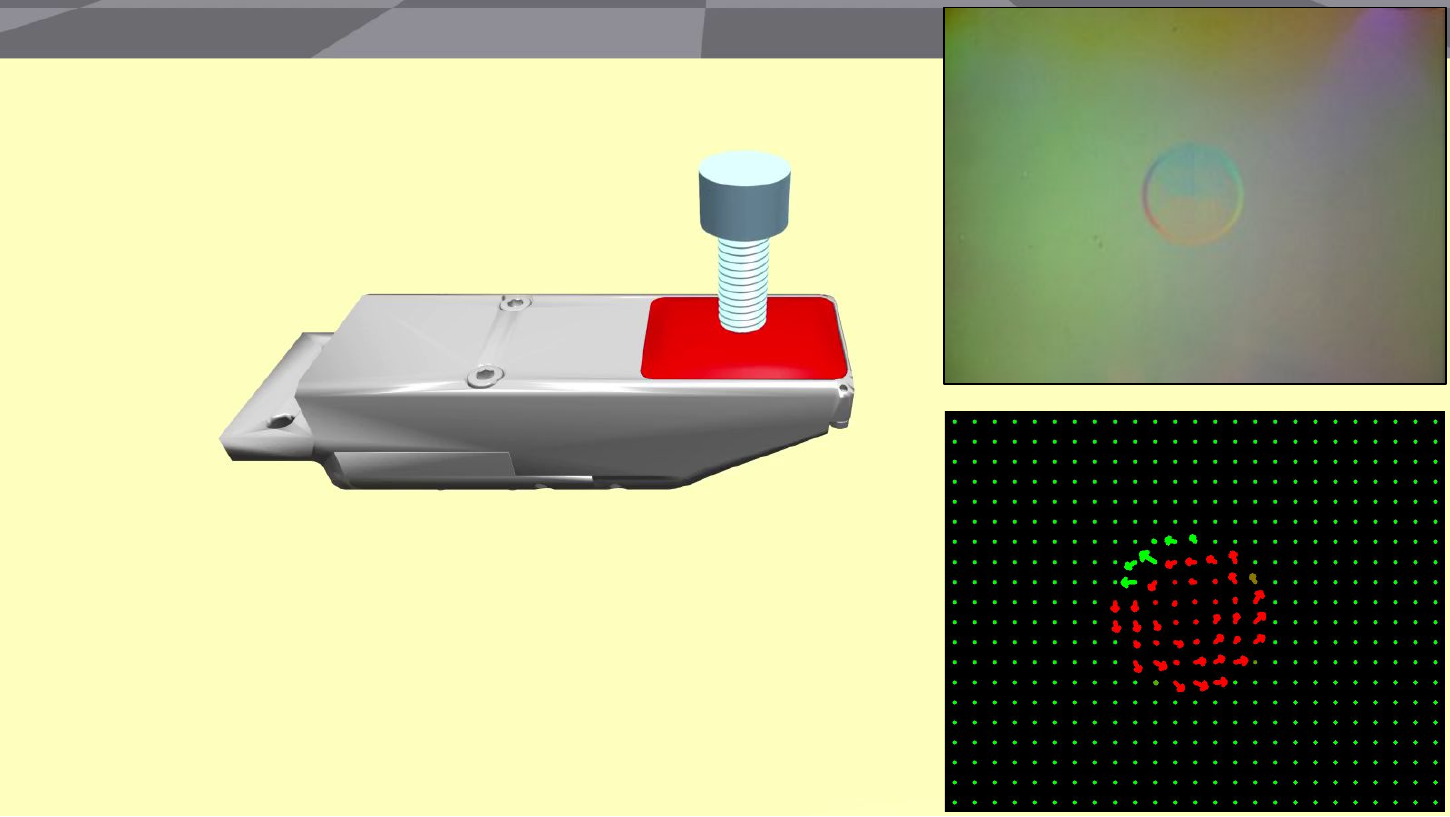} \\[\abovecaptionskip]
  \end{tabular}
  \begin{tabular}{@{}c@{}}
    \includegraphics[width=0.475\linewidth]{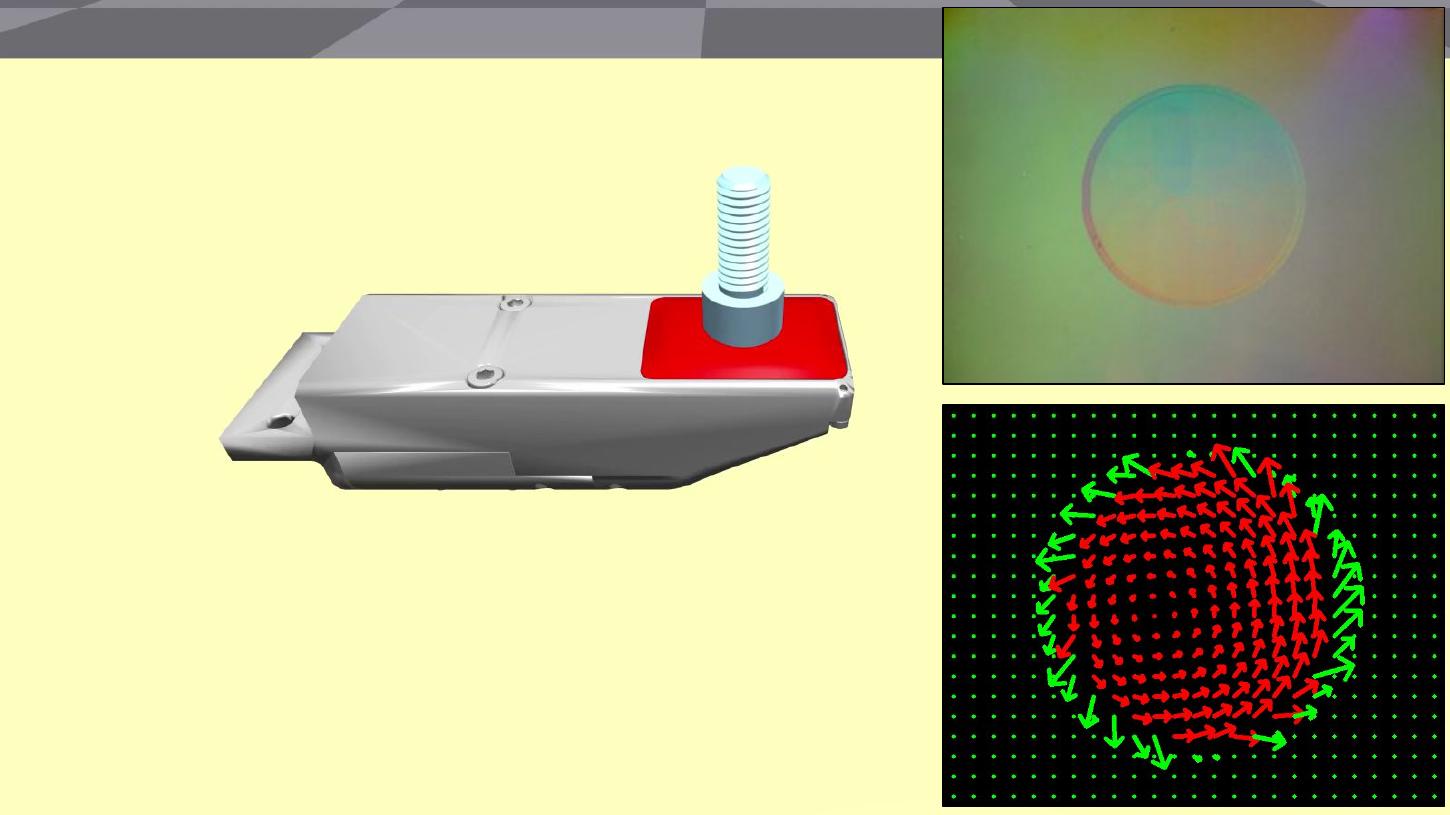} \\[\abovecaptionskip]
  \end{tabular}
\end{center}
\caption{
Additional visualizations of tactile readings (tactile image and shear force field) from the GelSight 1.5 sensor as the bolt mesh (30k faces and 26k vertices) interacts with the sensor. The bolt, positioned at varying poses, applies a downward force and rotates in the anticlockwise direction.
}
\label{fig:additional_viz}  
\end{figure}

\bibliographystyle{IEEEtran}
\bibliography{IEEEabrv,./references}

\begin{thebibliography}{10}
\providecommand{\url}[1]{#1}
\csname url@samestyle\endcsname
\providecommand{\newblock}{\relax}
\providecommand{\bibinfo}[2]{#2}
\providecommand{\BIBentrySTDinterwordspacing}{\spaceskip=0pt\relax}
\providecommand{\BIBentryALTinterwordstretchfactor}{4}
\providecommand{\BIBentryALTinterwordspacing}{\spaceskip=\fontdimen2\font plus
\BIBentryALTinterwordstretchfactor\fontdimen3\font minus \fontdimen4\font\relax}
\providecommand{\BIBforeignlanguage}[2]{{%
\expandafter\ifx\csname l@#1\endcsname\relax
\typeout{** WARNING: IEEEtran.bst: No hyphenation pattern has been}%
\typeout{** loaded for the language `#1'. Using the pattern for}%
\typeout{** the default language instead.}%
\else
\language=\csname l@#1\endcsname
\fi
#2}}
\providecommand{\BIBdecl}{\relax}
\BIBdecl

\bibitem{johansson2009coding}
R.~S. Johansson and J.~R. Flanagan, ``Coding and use of tactile signals from the fingertips in object manipulation tasks,'' \emph{Nature Reviews Neuroscience}, vol.~10, no.~5, pp. 345--359, 2009.

\bibitem{rogers2001passive}
M.~W. Rogers, D.~L. Wardman, S.~R. Lord, and R.~C. Fitzpatrick, ``Passive tactile sensory input improves stability during standing,'' \emph{Experimental Brain Research}, vol. 136, pp. 514--522, 2001.

\bibitem{ryan2021interaction}
C.~P. Ryan, G.~C. Bettelani, S.~Ciotti, C.~Parise, A.~Moscatelli, and M.~Bianchi, ``The interaction between motion and texture in the sense of touch,'' \emph{Journal of Neurophysiology}, vol. 126, no.~4, pp. 1375--1390, 2021.

\bibitem{luo2017robotic}
S.~Luo, J.~Bimbo, R.~Dahiya, and H.~Liu, ``Robotic tactile perception of object properties: A review,'' \emph{Mechatronics}, vol.~48, pp. 54--67, 2017.

\bibitem{li2020review}
Q.~Li, O.~Kroemer, Z.~Su, F.~F. Veiga, M.~Kaboli, and H.~J. Ritter, ``A review of tactile information: Perception and action through touch,'' \emph{IEEE Transactions on Robotics}, vol.~36, no.~6, pp. 1619--1634, 2020.

\bibitem{lepora2021soft}
N.~F. Lepora, ``Soft biomimetic optical tactile sensing with the tactip: A review,'' \emph{IEEE Sensors Journal}, vol.~21, no.~19, pp. 21\,131--21\,143, 2021.

\bibitem{she2021cable}
Y.~She, S.~Wang, S.~Dong, N.~Sunil, A.~Rodriguez, and E.~Adelson, ``Cable manipulation with a tactile-reactive gripper,'' \emph{The International Journal of Robotics Research}, vol.~40, no. 12-14, pp. 1385--1401, 2021.

\bibitem{zhao2023skill}
Y.~Zhao, X.~Jing, K.~Qian, D.~F. Gomes, and S.~Luo, ``Skill generalization of tubular object manipulation with tactile sensing and sim2real learning,'' \emph{Robotics and Autonomous Systems}, vol. 160, p. 104321, 2023.

\bibitem{yuan2017gelsight}
W.~Yuan, S.~Dong, and E.~H. Adelson, ``Gel{S}ight: High-resolution robot tactile sensors for estimating geometry and force,'' \emph{Sensors}, vol.~17, no.~12, p. 2762, 2017.

\bibitem{ward2018tactip}
B.~Ward-Cherrier, N.~Pestell, L.~Cramphorn, B.~Winstone, M.~E. Giannaccini, J.~Rossiter, and N.~F. Lepora, ``The {T}ac{T}ip family: Soft optical tactile sensors with 3{D}-printed biomimetic morphologies,'' \emph{Soft Robotics}, vol.~5, no.~2, pp. 216--227, 2018.

\bibitem{alspach2019soft}
A.~Alspach, K.~Hashimoto, N.~Kuppuswamy, and R.~Tedrake, ``Soft-bubble: A highly compliant dense geometry tactile sensor for robot manipulation,'' in \emph{2019 2nd IEEE International Conference on Soft Robotics (RoboSoft)}.\hskip 1em plus 0.5em minus 0.4em\relax IEEE, 2019, pp. 597--604.

\bibitem{lambeta2020digit}
M.~Lambeta, P.-W. Chou, S.~Tian, B.~Yang, B.~Maloon, V.~R. Most, D.~Stroud, R.~Santos, A.~Byagowi, G.~Kammerer \emph{et~al.}, ``Digit: A novel design for a low-cost compact high-resolution tactile sensor with application to in-hand manipulation,'' \emph{IEEE Robotics and Automation Letters}, vol.~5, no.~3, pp. 3838--3845, 2020.

\bibitem{makoviychuk2021isaac}
V.~Makoviychuk, L.~Wawrzyniak, Y.~Guo, M.~Lu, K.~Storey, M.~Macklin, D.~Hoeller, N.~Rudin, A.~Allshire, A.~Handa \emph{et~al.}, ``Isaac {G}ym: High performance {GPU}-based physics simulation for robot learning,'' \emph{arXiv preprint arXiv:2108.10470}, 2021.

\bibitem{todorov2012mujoco}
E.~Todorov, T.~Erez, and Y.~Tassa, ``Mu{J}o{C}o: A physics engine for model-based control,'' in \emph{2012 IEEE/RSJ international conference on intelligent robots and systems}.\hskip 1em plus 0.5em minus 0.4em\relax IEEE, 2012, pp. 5026--5033.

\bibitem{coumans2021}
E.~Coumans and Y.~Bai, ``Py{B}ullet: A {P}ython module for physics simulation for games, robotics and machine learning,'' \url{http://pybullet.org}, 2016--2021.

\bibitem{xu2023efficient}
J.~Xu, S.~Kim, T.~Chen, A.~R. Garcia, P.~Agrawal, W.~Matusik, and S.~Sueda, ``Efficient tactile simulation with differentiability for robotic manipulation,'' in \emph{Conference on Robot Learning}.\hskip 1em plus 0.5em minus 0.4em\relax PMLR, 2023, pp. 1488--1498.

\bibitem{narang2021sim}
Y.~Narang, B.~Sundaralingam, M.~Macklin, A.~Mousavian, and D.~Fox, ``Sim-to-real for robotic tactile sensing via physics-based simulation and learned latent projections,'' in \emph{2021 IEEE International Conference on Robotics and Automation (ICRA)}.\hskip 1em plus 0.5em minus 0.4em\relax IEEE, 2021, pp. 6444--6451.

\bibitem{si2022taxim}
Z.~Si and W.~Yuan, ``Taxim: An example-based simulation model for gelsight tactile sensors,'' \emph{IEEE Robotics and Automation Letters}, vol.~7, no.~2, pp. 2361--2368, 2022.

\bibitem{wang2022tacto}
S.~Wang, M.~Lambeta, P.-W. Chou, and R.~Calandra, ``Tacto: A fast, flexible, and open-source simulator for high-resolution vision-based tactile sensors,'' \emph{IEEE Robotics and Automation Letters}, vol.~7, no.~2, pp. 3930--3937, 2022.

\bibitem{lin2022tactile}
Y.~Lin, J.~Lloyd, A.~Church, and N.~F. Lepora, ``Tactile {G}ym 2.0: Sim-to-real deep reinforcement learning for comparing low-cost high-resolution robot touch,'' \emph{IEEE Robotics and Automation Letters}, vol.~7, no.~4, pp. 10\,754--10\,761, 2022.

\bibitem{hwangbo2019learning}
J.~Hwangbo, J.~Lee, A.~Dosovitskiy, D.~Bellicoso, V.~Tsounis, V.~Koltun, and M.~Hutter, ``Learning agile and dynamic motor skills for legged robots,'' \emph{Science Robotics}, vol.~4, no.~26, p. eaau5872, 2019.

\bibitem{chen2022system}
T.~Chen, J.~Xu, and P.~Agrawal, ``A system for general in-hand object re-orientation,'' in \emph{Conference on Robot Learning}.\hskip 1em plus 0.5em minus 0.4em\relax PMLR, 2022, pp. 297--307.

\bibitem{wettels2008biomimetic}
N.~Wettels, V.~J. Santos, R.~S. Johansson, and G.~E. Loeb, ``Biomimetic tactile sensor array,'' \emph{Advanced robotics}, vol.~22, no.~8, pp. 829--849, 2008.

\bibitem{bhirangi2021reskin}
R.~Bhirangi, T.~Hellebrekers, C.~Majidi, and A.~Gupta, ``Reskin: versatile, replaceable, lasting tactile skins,'' \emph{arXiv preprint arXiv:2111.00071}, 2021.

\bibitem{johnson2009retrographic}
M.~K. Johnson and E.~H. Adelson, ``Retrographic sensing for the measurement of surface texture and shape,'' in \emph{2009 IEEE Conference on Computer Vision and Pattern Recognition}.\hskip 1em plus 0.5em minus 0.4em\relax IEEE, 2009, pp. 1070--1077.

\bibitem{Reddy2019Book}
J.~N. Reddy, \emph{Introduction to the Finite Element Method}.\hskip 1em plus 0.5em minus 0.4em\relax New York, USA: McGraw Hill Education, 2019.

\bibitem{narang2021interpreting}
Y.~S. Narang, B.~Sundaralingam, K.~Van~Wyk, A.~Mousavian, and D.~Fox, ``Interpreting and predicting tactile signals for the syntouch biotac,'' \emph{The International Journal of Robotics Research}, vol.~40, no. 12-14, pp. 1467--1487, 2021.

\bibitem{ma2019dense}
D.~Ma, E.~Donlon, S.~Dong, and A.~Rodriguez, ``Dense tactile force estimation using {G}el{S}lim and inverse {FEM},'' in \emph{2019 International Conference on Robotics and Automation (ICRA)}.\hskip 1em plus 0.5em minus 0.4em\relax IEEE, 2019, pp. 5418--5424.

\bibitem{sferrazza2019ground}
C.~Sferrazza, A.~Wahlsten, C.~Trueeb, and R.~D’Andrea, ``Ground truth force distribution for learning-based tactile sensing: A finite element approach,'' \emph{IEEE Access}, vol.~7, pp. 173\,438--173\,449, 2019.

\bibitem{sferrazza2020learning}
C.~Sferrazza, T.~Bi, and R.~D’Andrea, ``Learning the sense of touch in simulation: a sim-to-real strategy for vision-based tactile sensing,'' in \emph{2020 IEEE/RSJ International Conference on Intelligent Robots and Systems (IROS)}.\hskip 1em plus 0.5em minus 0.4em\relax IEEE, 2020, pp. 4389--4396.

\bibitem{huang2022defgraspsim}
I.~Huang, Y.~Narang, C.~Eppner, B.~Sundaralingam, M.~Macklin, R.~Bajcsy, T.~Hermans, and D.~Fox, ``Def{G}rasp{S}im: Physics-based simulation of grasp outcomes for 3{D} deformable objects,'' \emph{IEEE Robotics and Automation Letters}, vol.~7, no.~3, pp. 6274--6281, 2022.

\bibitem{pfaff2020learning}
T.~Pfaff, M.~Fortunato, A.~Sanchez-Gonzalez, and P.~W. Battaglia, ``Learning mesh-based simulation with graph networks,'' \emph{arXiv preprint arXiv:2010.03409}, 2020.

\bibitem{huang2023defgraspnets}
I.~Huang, Y.~Narang, R.~Bajcsy, F.~Ramos, T.~Hermans, and D.~Fox, ``Def{G}rasp{N}ets: Grasp planning on 3d fields with graph neural nets,'' \emph{International Conference on Robotics and Automation (ICRA)}, 2023.

\bibitem{10388459}
W.~Chen, J.~Xu, F.~Xiang, X.~Yuan, H.~Su, and R.~Chen, ``General-purpose sim2real protocol for learning contact-rich manipulation with marker-based visuotactile sensors,'' \emph{IEEE Transactions on Robotics}, vol.~40, pp. 1509--1526, 2024.

\bibitem{elandt2019pressure}
R.~Elandt, E.~Drumwright, M.~Sherman, and A.~Ruina, ``A pressure field model for fast, robust approximation of net contact force and moment between nominally rigid objects,'' in \emph{2019 IEEE/RSJ International Conference on Intelligent Robots and Systems (IROS)}.\hskip 1em plus 0.5em minus 0.4em\relax IEEE, 2019, pp. 8238--8245.

\bibitem{church2022tactile}
A.~Church, J.~Lloyd, N.~F. Lepora \emph{et~al.}, ``Tactile sim-to-real policy transfer via real-to-sim image translation,'' in \emph{Conference on Robot Learning}.\hskip 1em plus 0.5em minus 0.4em\relax PMLR, 2022, pp. 1645--1654.

\bibitem{gomes2021generation}
D.~F. Gomes, P.~Paoletti, and S.~Luo, ``Generation of {G}el{S}ight tactile images for sim2real learning,'' \emph{IEEE Robotics and Automation Letters}, vol.~6, no.~2, pp. 4177--4184, 2021.

\bibitem{jing2023unsupervised}
X.~Jing, K.~Qian, T.~Jianu, and S.~Luo, ``Unsupervised adversarial domain adaptation for sim-to-real transfer of tactile images,'' \emph{IEEE Transactions on Instrumentation and Measurement}, 2023.

\bibitem{chen2022bidirectional}
W.~Chen, Y.~Xu, Z.~Chen, P.~Zeng, R.~Dang, R.~Chen, and J.~Xu, ``Bidirectional sim-to-real transfer for {G}el{S}ight tactile sensors with {C}ycle{GAN},'' \emph{IEEE Robotics and Automation Letters}, vol.~7, no.~3, pp. 6187--6194, 2022.

\bibitem{higuera2023learning}
C.~Higuera, B.~Boots, and M.~Mukadam, ``Learning to read braille: Bridging the tactile reality gap with diffusion models,'' \emph{arXiv preprint arXiv:2304.01182}, 2023.

\bibitem{ding2020sim}
Z.~Ding, N.~F. Lepora, and E.~Johns, ``Sim-to-real transfer for optical tactile sensing,'' in \emph{2020 IEEE International Conference on Robotics and Automation (ICRA)}.\hskip 1em plus 0.5em minus 0.4em\relax IEEE, 2020, pp. 1639--1645.

\bibitem{do2023densetact}
W.~K. Do, A.~K. Dhawan, M.~Kitzmann, and M.~Kennedy~III, ``Densetact-mini: An optical tactile sensor for grasping multi-scale objects from flat surfaces,'' \emph{arXiv preprint arXiv:2309.08860}, 2023.

\bibitem{lin2023bi}
Y.~Lin, A.~Church, M.~Yang, H.~Li, J.~Lloyd, D.~Zhang, and N.~F. Lepora, ``Bi-touch: Bimanual tactile manipulation with sim-to-real deep reinforcement learning,'' \emph{IEEE Robotics and Automation Letters}, 2023.

\bibitem{bi2021zero}
T.~Bi, C.~Sferrazza, and R.~D’Andrea, ``Zero-shot sim-to-real transfer of tactile control policies for aggressive swing-up manipulation,'' \emph{IEEE Robotics and Automation Letters}, vol.~6, no.~3, pp. 5761--5768, 2021.

\bibitem{qi2023general}
H.~Qi, B.~Yi, S.~Suresh, M.~Lambeta, Y.~Ma, R.~Calandra, and J.~Malik, ``General in-hand object rotation with vision and touch,'' \emph{arXiv preprint arXiv:2309.09979}, 2023.

\bibitem{PhysX}
\BIBentryALTinterwordspacing
{NVIDIA Corporation}, ``Nvidia physx sdk,'' 2024, version 5.4. [Online]. Available: \url{https://github.com/NVIDIA-Omniverse/PhysX}
\BIBentrySTDinterwordspacing

\bibitem{macklin2019small}
M.~Macklin, K.~Storey, M.~Lu, P.~Terdiman, N.~Chentanez, S.~Jeschke, and M.~M{\"u}ller, ``Small steps in physics simulation,'' in \emph{Proceedings of the 18th annual ACM siggraph/eurographics symposium on computer animation}, 2019, pp. 1--7.

\bibitem{fliigge1967viscoelasticity}
W.~Fliigge, ``Viscoelasticity,'' \emph{Blaisdell Publ. Comp., London}, pp. 1069--1084, 1967.

\bibitem{kao2016contact}
I.~Kao, K.~M. Lynch, and J.~W. Burdick, ``Contact modeling and manipulation,'' \emph{Springer Handbook of Robotics}, pp. 931--954, 2016.

\bibitem{tan2011stable}
J.~Tan, K.~Liu, and G.~Turk, ``Stable proportional-derivative controllers,'' \emph{IEEE Computer Graphics and Applications}, vol.~31, no.~4, pp. 34--44, 2011.

\bibitem{xu2021end}
J.~Xu, T.~Chen, L.~Zlokapa, M.~Foshey, W.~Matusik, S.~Sueda, and P.~Agrawal, ``An end-to-end differentiable framework for contact-aware robot design,'' \emph{arXiv preprint arXiv:2107.07501}, 2021.

\bibitem{narang2022factory}
Y.~Narang, K.~Storey, I.~Akinola, M.~Macklin, P.~Reist, L.~Wawrzyniak, Y.~Guo, A.~Moravanszky, G.~State, M.~Lu \emph{et~al.}, ``Factory: Fast contact for robotic assembly,'' \emph{Robotics: Science and Systems}, 2022.

\bibitem{macklin2020local}
M.~Macklin, K.~Erleben, M.~M{\"u}ller, N.~Chentanez, S.~Jeschke, and Z.~Corse, ``Local optimization for robust signed distance field collision,'' \emph{Proceedings of the ACM on Computer Graphics and Interactive Techniques}, vol.~3, no.~1, pp. 1--17, 2020.

\bibitem{torabi2018behavioral}
F.~Torabi, G.~Warnell, and P.~Stone, ``Behavioral cloning from observation,'' \emph{arXiv preprint arXiv:1805.01954}, 2018.

\bibitem{ross2011reduction}
S.~Ross, G.~Gordon, and D.~Bagnell, ``A reduction of imitation learning and structured prediction to no-regret online learning,'' in \emph{Proceedings of the fourteenth international conference on artificial intelligence and statistics}.\hskip 1em plus 0.5em minus 0.4em\relax JMLR Workshop and Conference Proceedings, 2011, pp. 627--635.

\bibitem{rusu2015policy}
A.~A. Rusu, S.~G. Colmenarejo, C.~Gulcehre, G.~Desjardins, J.~Kirkpatrick, R.~Pascanu, V.~Mnih, K.~Kavukcuoglu, and R.~Hadsell, ``Policy distillation,'' \emph{arXiv preprint arXiv:1511.06295}, 2015.

\bibitem{czarnecki2019distilling}
W.~M. Czarnecki, R.~Pascanu, S.~Osindero, S.~Jayakumar, G.~Swirszcz, and M.~Jaderberg, ``Distilling policy distillation,'' in \emph{The 22nd international conference on artificial intelligence and statistics}.\hskip 1em plus 0.5em minus 0.4em\relax PMLR, 2019, pp. 1331--1340.

\bibitem{schulman2017proximal}
J.~Schulman, F.~Wolski, P.~Dhariwal, A.~Radford, and O.~Klimov, ``Proximal policy optimization algorithms,'' \emph{arXiv preprint arXiv:1707.06347}, 2017.

\bibitem{pinto2017asymmetric}
L.~Pinto, M.~Andrychowicz, P.~Welinder, W.~Zaremba, and P.~Abbeel, ``Asymmetric actor critic for image-based robot learning,'' \emph{arXiv preprint arXiv:1710.06542}, 2017.

\bibitem{wang2021gelsight}
S.~Wang, Y.~She, B.~Romero, and E.~Adelson, ``Gel{S}ight {W}edge: Measuring high-resolution 3{D} contact geometry with a compact robot finger,'' in \emph{2021 IEEE International Conference on Robotics and Automation (ICRA)}.\hskip 1em plus 0.5em minus 0.4em\relax IEEE, 2021, pp. 6468--6475.

\bibitem{tang2024automate}
B.~Tang, I.~Akinola, J.~Xu, B.~Wen, A.~Handa, K.~Van~Wyk, D.~Fox, G.~S. Sukhatme, F.~Ramos, and Y.~Narang, ``Automate: Specialist and generalist assembly policies over diverse geometries,'' \emph{arXiv preprint arXiv:2407.08028}, 2024.

\bibitem{tang2023industreal}
B.~Tang, M.~A. Lin, I.~Akinola, A.~Handa, G.~S. Sukhatme, F.~Ramos, D.~Fox, and Y.~Narang, ``Indust{R}eal: Transferring contact-rich assembly tasks from simulation to reality,'' \emph{Robotics: Science and Systems}, 2023.

\bibitem{hunt1975coefficient}
K.~H. Hunt and F.~R.~E. Crossley, ``Coefficient of restitution interpreted as damping in vibroimpact,'' 1975.

\bibitem{marhefka1999compliant}
D.~W. Marhefka and D.~E. Orin, ``A compliant contact model with nonlinear damping for simulation of robotic systems,'' \emph{IEEE Transactions on Systems, Man, and Cybernetics-Part A: Systems and Humans}, vol.~29, no.~6, pp. 566--572, 1999.

\bibitem{gomes2020geltip}
D.~F. Gomes, Z.~Lin, and S.~Luo, ``Geltip: A finger-shaped optical tactile sensor for robotic manipulation,'' in \emph{2020 IEEE/RSJ International Conference on Intelligent Robots and Systems (IROS)}.\hskip 1em plus 0.5em minus 0.4em\relax IEEE, 2020, pp. 9903--9909.

\bibitem{tippur2023gelsight360}
M.~H. Tippur and E.~H. Adelson, ``Gelsight360: An omnidirectional camera-based tactile sensor for dexterous robotic manipulation,'' in \emph{2023 IEEE International Conference on Soft Robotics (RoboSoft)}.\hskip 1em plus 0.5em minus 0.4em\relax IEEE, 2023, pp. 1--8.

\bibitem{noseworthy2024forge}
M.~Noseworthy, B.~Tang, B.~Wen, A.~Handa, N.~Roy, D.~Fox, F.~Ramos, Y.~Narang, and I.~Akinola, ``Forge: Force-guided exploration for robust contact-rich manipulation under uncertainty,'' \emph{arXiv preprint arXiv:2408.04587}, 2024.

\end{thebibliography}

\vfill\eject    

\begin{IEEEbiography}[{\includegraphics[width=1in,height=1.25in,clip,keepaspectratio]{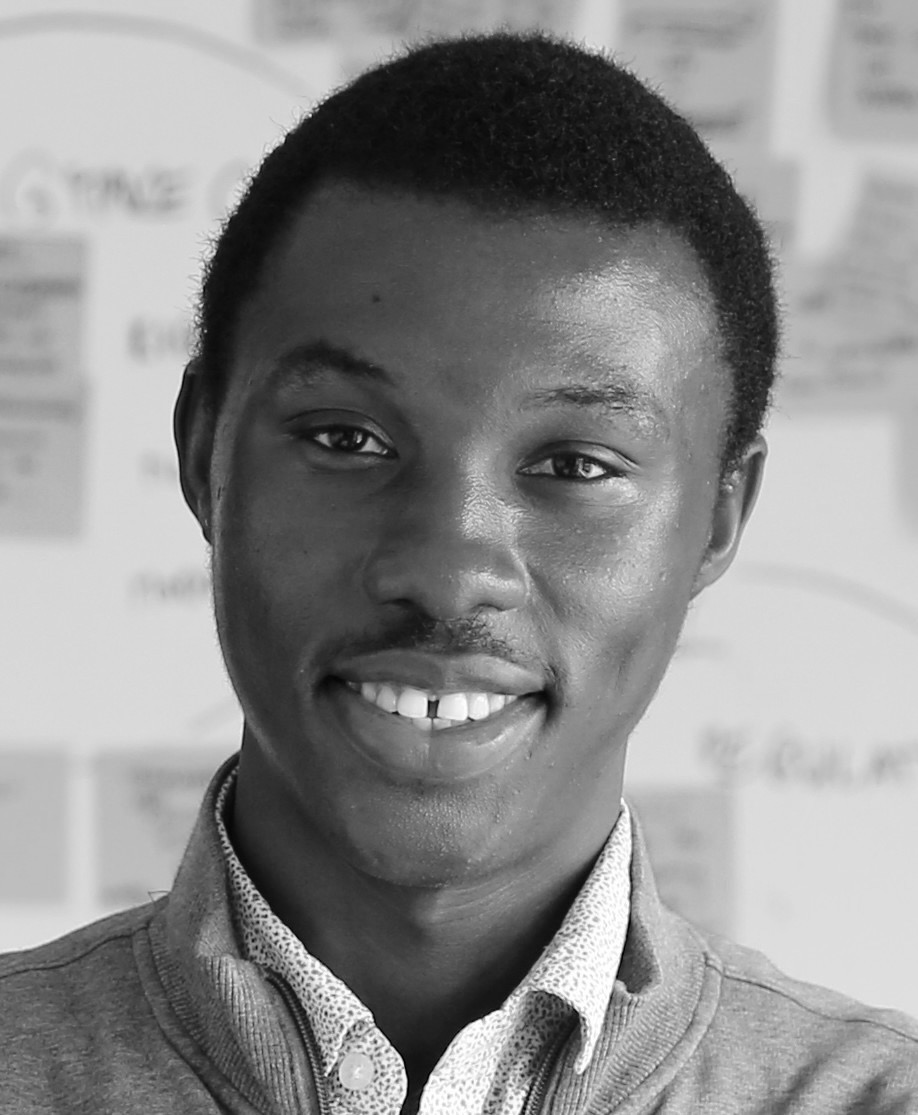}}]{Iretiayo Akinola}
Iretiayo Akinola received a PhD from Columbia University's Computer Science Department in 2021, where his research centered on equipping robots with multi-modal (visual and tactile) perception for reactive object grasping and manipulation. Prior to that, he earned his M.S. and BSc. degrees in Electrical Engineering from Stanford University and Obafemi Awolowo University, respectively. His research interests include multi-modal robot learning, sim-to-real transfer, and human-in-the-loop robot learning.
\end{IEEEbiography}
\vspace{-20pt}
\begin{IEEEbiography}[{\includegraphics[width=1in,height=1.25in,clip,keepaspectratio]{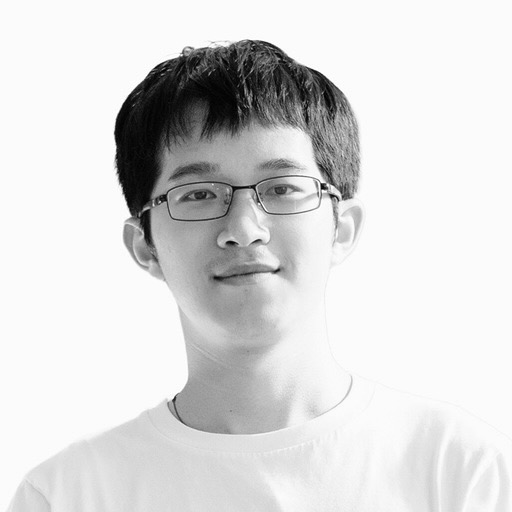}}]{Jie Xu}
Jie Xu received the B.Eng. degree from Department of  Computer Science and Technology at Tsinghua University in 2016 and Ph.D. degree in Computer Science from Massachusetts Institute of Technology in 2022. He is currently a Research Scientist at Nvidia Research. His research interests include Robotics Control, Reinforcement Learning, Simulation, Robot Co-Design and Sim-to-Real.
\end{IEEEbiography}
\vspace{-20pt}
\begin{IEEEbiography}[{\includegraphics[width=1in,height=1.25in,clip,keepaspectratio]{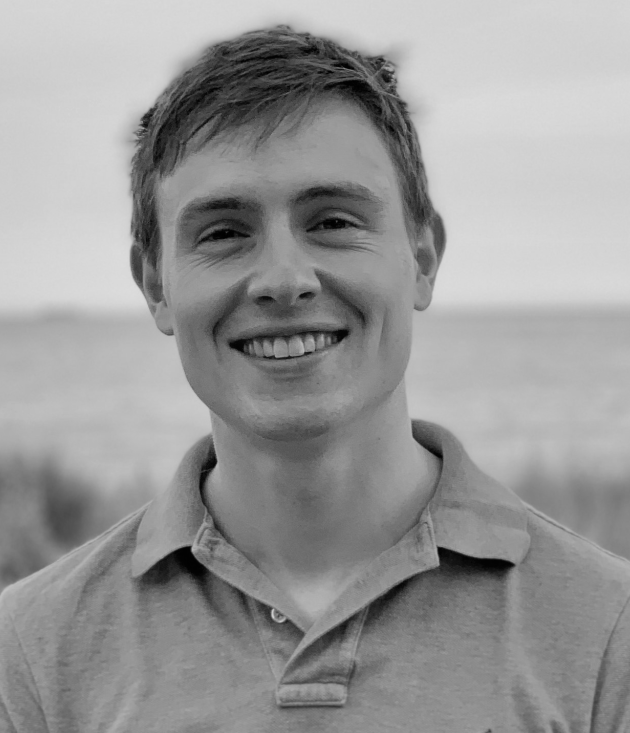}}]{Jan Carius}
Jan Carius is currently working in the domain of physics simulations and parallel computing.
He received a PhD from ETH Zurich in 2021 where his research evolved around motion planning and control algorithms for legged locomotion based on optimal control and machine learning.
Formerly he received BSc. and MSc. degrees in mechanical engineering from ETH Zurich.
\end{IEEEbiography}
\vspace{-10pt}
\begin{IEEEbiography}[{\includegraphics[width=1in,height=1.25in,clip,keepaspectratio]{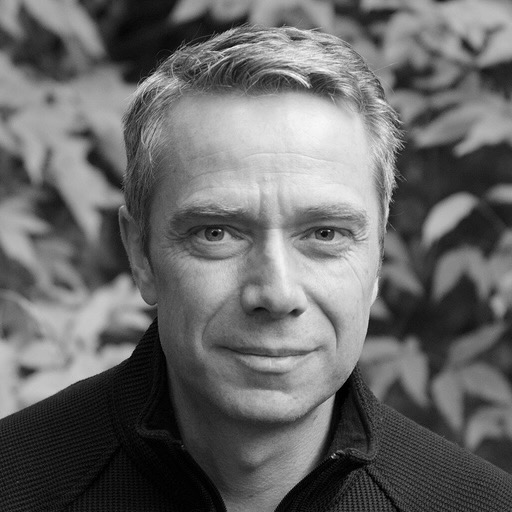}}]{Dieter Fox}
Dieter Fox received his PhD degree from the University of Bonn, Germany. He is a
professor in the Allen School of Computer Science \& Engineering at the
University of Washington, where he heads the UW Robotics and State Estimation
Lab.  He is also Senior Director of Robotics Research at NVIDIA.  His research
is in robotics and artificial intelligence, with a focus on learning and
estimation applied to problems such as robot manipulation, planning, language
grounding, and activity recognition. He has published more than 300 technical
papers and is co-author of the textbook “Probabilistic Robotics”. Dieter is a
Fellow of the IEEE, ACM, and AAAI, and recipient of the IEEE RAS Pioneer Award
and the IJCAI John McCarthy Award.
\end{IEEEbiography}
\vspace{-10pt}
\begin{IEEEbiography}[{\includegraphics[width=1in,height=1.25in,clip,keepaspectratio]{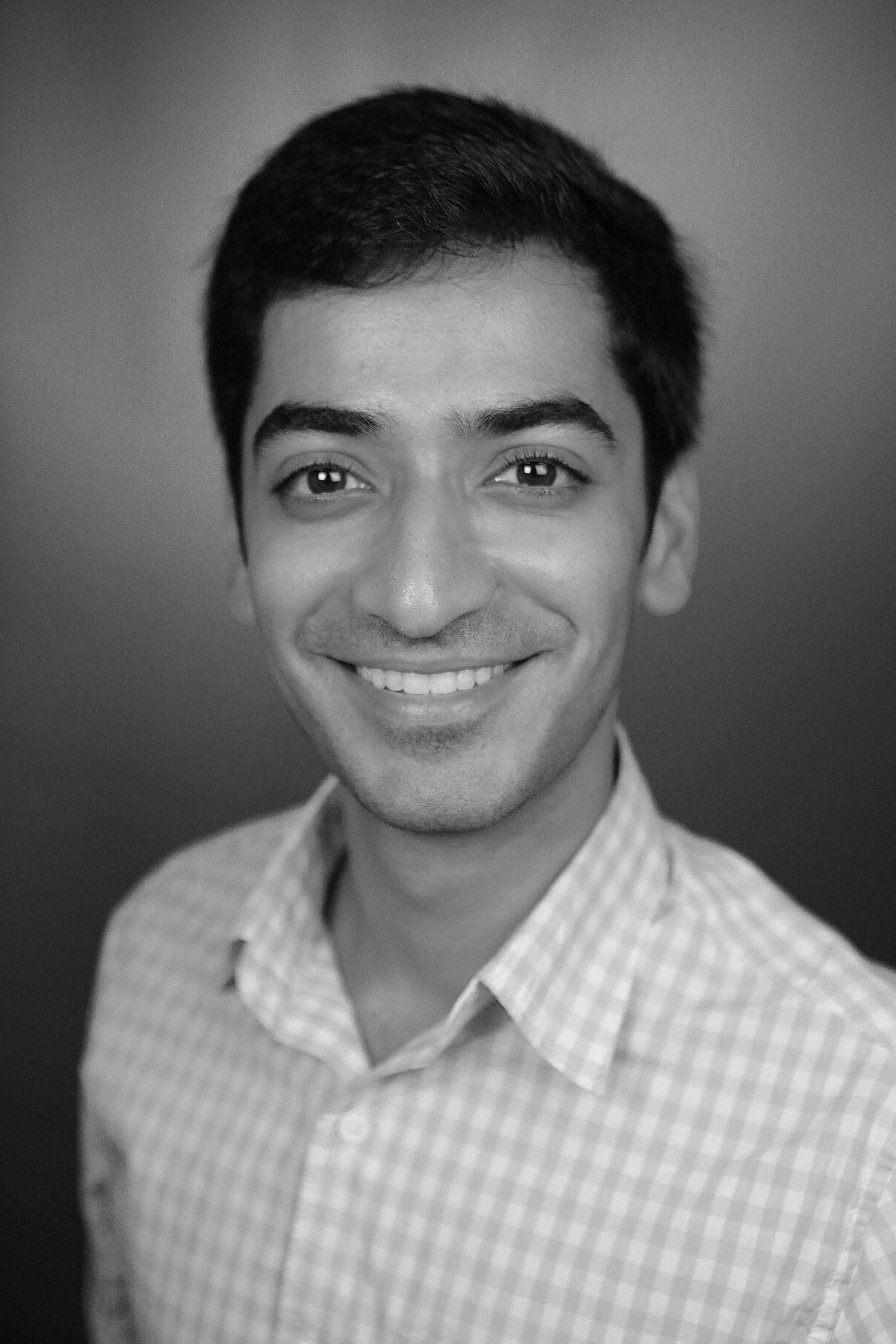}}]{Yashraj Narang}
Yashraj Narang received a B.S. from Stanford University in 2011, an S.M. from the Massachusetts Institute of Technology in 2013, and a Ph.D. from Harvard University in 2018. He joined the NVIDIA Seattle Robotics Lab as a research scientist in 2018 and has led the Simulation and Behavior Generation group since 2023. His current research focuses on leveraging physics-based simulation for robot learning in structured and unstructured environments.
\end{IEEEbiography}

\vfill

\end{document}